\documentclass{article}
\usepackage{ijcai25}

\usepackage{times}
\usepackage{soul}
\usepackage{url}
\usepackage[hidelinks]{hyperref}
\usepackage[utf8]{inputenc}
\usepackage[small]{caption}
\usepackage{graphicx}
\usepackage{amsmath}
\usepackage{amsthm}
\usepackage{booktabs}
\usepackage{algorithm}
\usepackage{algorithmic}

\usepackage[switch]{lineno}

\usepackage{multirow}
\usepackage{amsfonts}
\usepackage{colortbl}
\definecolor{lightgray}{gray}{0.9}
\definecolor{deepgreen}{rgb}{0,0.5,0}

\title{GKG-LLM: A Unified Framework for Generalized Knowledge Graph Construction}

\author{Jian Zhang$^{1,3}$, Bifan Wei$^{2,3}$, Shihao Qi$^{1,3}$, Haiping Zhu$^{1,3}$, Jun Liu$^{1,3}$, Qika Lin$^{4}$\thanks{Corresponding author}\\
\affiliations
	$^{1}$School of Computer Science and Technology, Xi'an Jiaotong University, Xi'an, China \\ 
    $^{2}$School of Continuing Education, Xi'an Jiaotong University, Xi'an, China\\
    $^{3}$Shaanxi Province Key Laboratory of Big Data Knowledge Engineering, \\Xi’an Jiaotong University, Xi'an, China \\
    $^{4}$National University of Singapore \\
    \emails
    \{zhangjian062422@stu., weibifan@, chihoq@stu., zhuhaiping@, liukeen@\}xjtu.edu.cn, linqika@nus.edu.sg}

\begin{document}

\maketitle

\begin{abstract}
    The construction of Generalized Knowledge Graph (GKG), including knowledge graph, event knowledge graph and commonsense knowledge graph, is fundamental for various natural language processing tasks. Current studies typically construct these types of graph separately, overlooking holistic insights and potential unification that could be beneficial in computing resources and usage perspectives. However, a key challenge in developing a unified framework for GKG is obstacles arising from task-specific differences. In this study, we propose a unified framework for constructing generalized knowledge graphs to address this challenge. First, we collect data from 15 sub-tasks in 29 datasets across the three types of graphs, categorizing them into in-sample, counter-task, and out-of-distribution (OOD) data. Then, we propose a three-stage curriculum learning fine-tuning framework, by iteratively injecting knowledge from the three types of graphs into the Large Language Models. Extensive experiments show that our proposed model improves the construction of all three graph types across in-domain, OOD and counter-task data.
\end{abstract}

\section{Introduction} \label{1}

\begin{figure}[t]
  \includegraphics[width=\columnwidth]{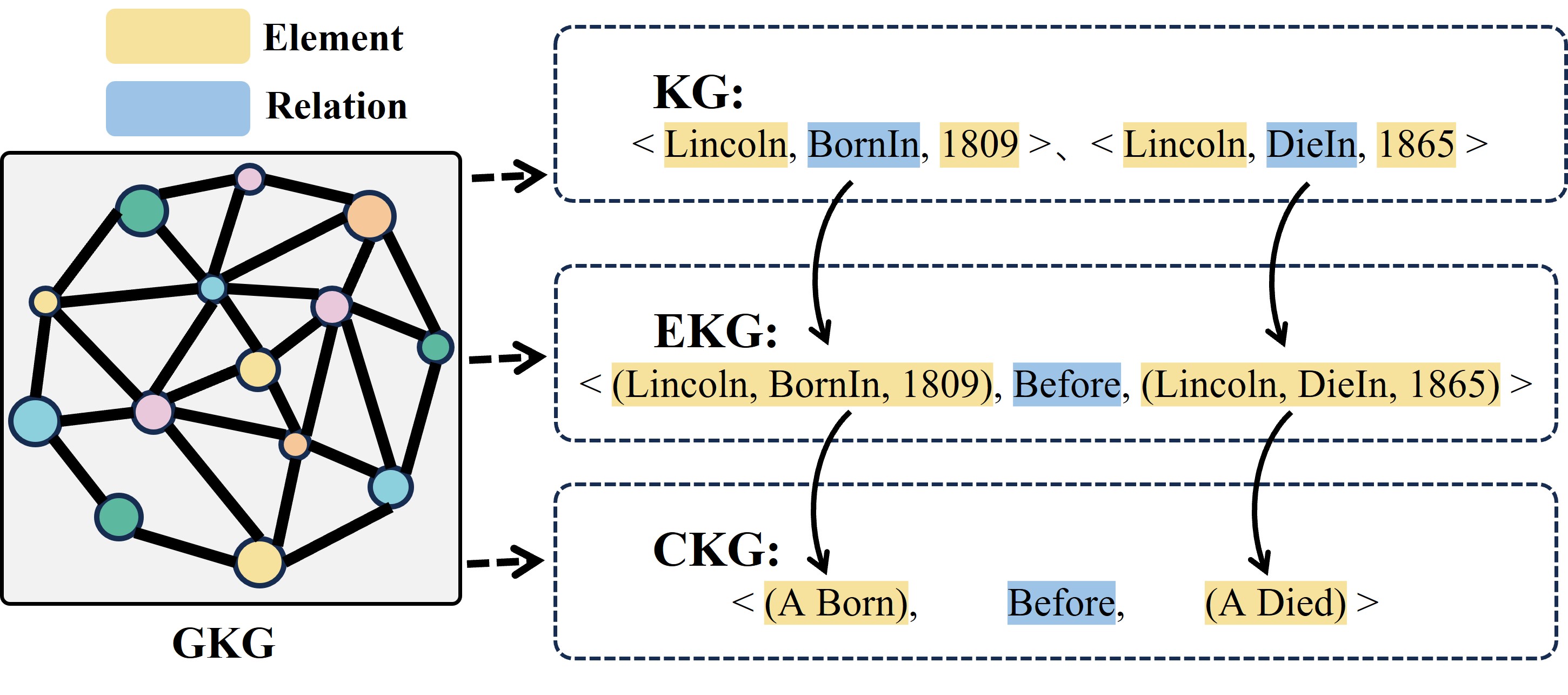}
  \caption{An illustration of several triples and graphs. The left half shows a generalized knowledge graph. The right half includes specific examples of triples from KG, EKG, CKG and demonstrates their progressive relationship.}
  \label{fig_example}
\end{figure}

Generalized Knowledge Graph (GKG) \cite{krause2022generalized} includes Knowledge Graph (KG), Event Knowledge Graph (EKG) and Commonsense Knowledge Graph (CKG). The construction of GKG encompasses multiple essential tasks \cite{peng2023knowledge}, which are crucial for various applications in this field, including intelligence analysis \cite{pimenov2023artificial} and decision support \cite{lai2023towards}. As shown in Figure~\ref{fig_example}, KGs \cite{DBLP:conf/acl/LinL0XC23,DBLP:journals/corr/abs-2501-18119}
are developed to more effectively describe concepts and relations in the physical world. The fundamental structure is \textit{ \textless entity, relation, entity \textgreater }, such as \textless Lincoln, BornIn, 1809\textgreater. With ongoing research, EKGs are introduced to study the dynamic progression of events. It is organized in the triplet format \textit{ \textless event, relation, event \textgreater }, as illustrated by \textless (Lincoln, BornIn, 1809), Before, (Lincoln, diedIn, 1865) \textgreater . The further generalization of event graphs has led to the development of CKG, which abstractly represent general relational patterns in the form of \textit{ \textless commonsense, relation, commonsense\textgreater }. For instance, \textless (A born), Before, (A died)\textgreater   is also organized in a triplet format. In summary, KG, EKG, and CKG are all organized in the basic form of \textbf{ \textless element, relation, element \textgreater }.  

Overall, constructing the three types of graphs separately requires substantial resources, while using a unified framework for their construction improves parameter efficiency. Additionally, from a usage perspective, the knowledge contained in KGs facilitates the construction of both EKGs and CKGs. For example, a method leveraging hierarchical KGs to enhance the accuracy and effectiveness of biomedical event extraction is proposed by \cite{huang2020biomedical}. Similarly, for knowledge graphs aiding text classification in the construction of CKGs, KG-MTT-BERT \cite{he2022kg} is introduced to enhance BERT with KGs for multi-type medical text classification.

\begin{figure}[t]
    \includegraphics[width=\columnwidth]{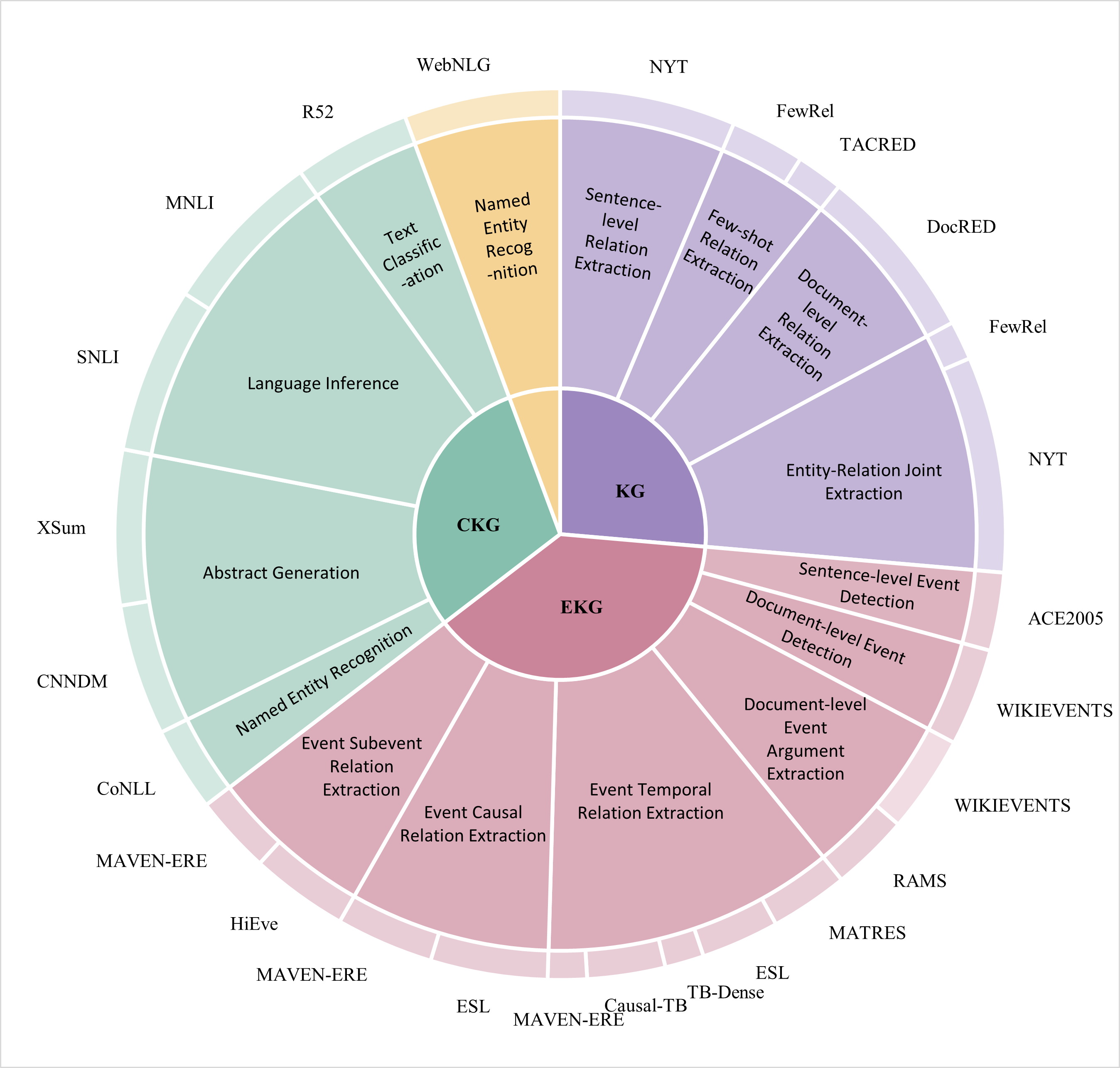}
    \caption{The illustration of the data distribution for all GKG sub-tasks.}
    \label{data_distribution}
\end{figure}

Naturally, we abstract a new task to build a unified framework for constructing GKG, in order to empower these foundational triples extraction tasks. However, a key challenge in this task is the obstacles arising from task-specific differences.
The construction of different types of graph involves a wide variety of diverse sub-tasks. Specifically, as illustrated in Figure~\ref{data_distribution}, the construction of KG includes sub-tasks such as sentence-level relation extraction \cite{wadhwa2023revisiting}, document-level relation extraction \cite{ma2023dreeam} and joint entity and relation extraction \cite{sui2023joint}. The construction of EKG involves sub-tasks such as sentence-level event detection \cite{hettiarachchi2023ttl}, document-level argument extraction 
 \cite{zhang2024semantic}, and event temporal relation extraction \cite{chan2024exploring}. While the construction of CKG includes sub-tasks such as abstract generation \cite{gao2023comparing} and language inference \cite{gubelmann2024capturing}. The abbreviations and introduction of the task can be found in Appendix~\ref{sub-tasks}.
These tasks differ in several ways, with the primary distinctions lying in their definitions and content. For instance, sentence-level relation extraction involves extracting the relationship between two entities from a single sentence, whereas abstract generation involves extracting an abstract from an entire article. Differences between these tasks have created obstacles to building a unified framework for constructing GKG.

Thanks to the emergence of Large Language Models(LLMs), such as GPT4 \cite{achiam2023gpt} and LlaMA-3 \cite{dubey2024llama}, the realization of this new unified task has become possible. The standardized input-output format of LLMs unifies these sub-tasks from a structural perspective. To this end, we propose a three-stage curriculum learning tuning framework. Firstly, data collection and preparation involve extensively gathering data from three types of graphs, resulting in a total of 15 sub-tasks in 29 datasets. These datasets are categorized into three types: conventional datasets for training and testing, counter-task datasets also used for training and testing to prevent model overfitting and enhance generalization, and out-of-distribution (OOD) datasets used solely for testing. Secondly, the three-stage curriculum learning fine-tuning framework, built upon a base model, includes the \textit{KG Empowerment Stage}, which leverages KG datasets, the \textit{EKG Enhancement Stage}, utilizing EKG datasets, and the \textit{CKG Generalization Stage}, which incorporates CKG datasets along with counter-task datasets. Through these three stages of training, we obtain the micro, mid, and macro versions of GKG-LLM, respectively. Finally, GKG-LLM has undergone extensive testing and analysis on all three graph types across in-domain, OOD, and counter-task data, demonstrating the effectiveness and advancement of diverse instruction design strategies and the three-stage fine-tuning framework.

\begin{figure*}[t]
	\large
	\centering

	\includegraphics[scale=0.35]{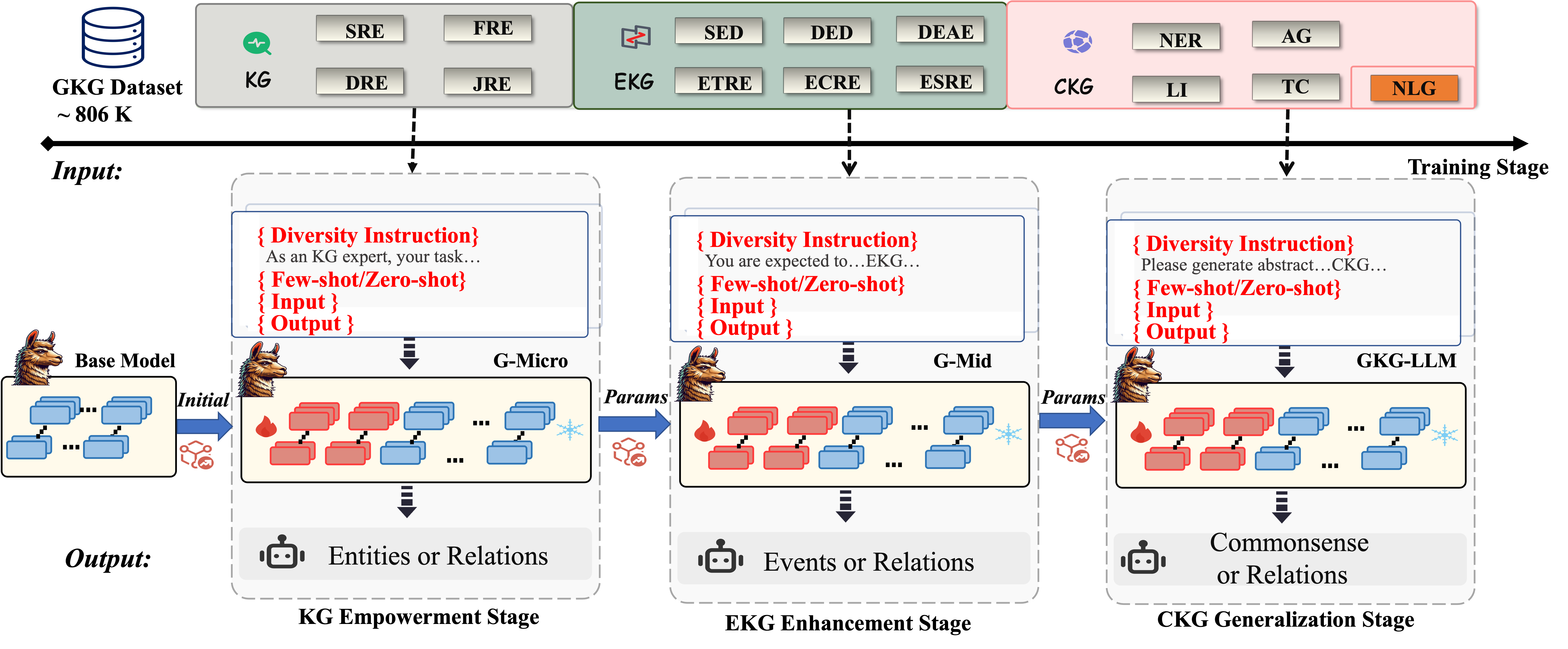}
	\caption{Three-stage curriculum learning tuning framework of GKG-LLM. The upper part represents the GKG dataset \( \mathcal{D}_G\), consisting of the unified datasets. The lower part shows the three stages of GKG training: the \textit{KG empowerment stage} using the KG datasets to build foundational skills, the \textit{EKG enhancement stage} using the EKG datasets to enhance specific capabilities, and the \textit{CKG generalization stage} using the CKG datasets and the counter task dataset to achieve generalization of the GKG-LLM capabilities. The thick arrows between the stages represent the delivery of model parameters from base model to each version of GKG-LLM.}
	\label{fig_model}
	\vspace{-0.3cm}
\end{figure*}

The contributions of this research are listed as follows:
\begin{itemize}
    \item{We propose an approach for building GKG using a three-stage curriculum learning fine-tuning framework, resulting in a GKG-LLM\footnote{https://anonymous.4open.science/r/GKG-sample-64DB. This part is the core weight of the code. Once finalized, the manuscript will be shared with the open-source community.} that addresses task-specific differences and enables the unified construction of GKG.}
    \item{From a data perspective, this study is the first to collect and process sub-task datasets from three types of graphs in a comprehensive view, exploring their intrinsic connections in constructing GKG, as far as we know.}
    \item {Extensive experiments report that GKG-LLM achieves the effectiveness and advancement on three types of data and further analysis validates the superiority of our architecture.}
\end{itemize}

\section{Methodology}

In this section, we first present the three-stage curriculum learning tuning framework in Section~\ref{2.1}, then describe data collection and preparation in Section~\ref{2.2} and introduce our training strategy in Section~\ref{2.3}.

The formal definition of GKG construction involves reformulating the various sub-tasks of KG, EKG, and CKG using a unified seq2seq format and structure. Then we solve it through three-stage fine-tuning LLMs, as shown in Figure~\ref{fig_model}. Specifically, the unified input is a task document or sentence, and the unified output consists of the elements or relations that form the GKG triples.

\subsection{GKG-LLM} \label{2.1}

The overview of GKG-LLM is shown in Figure~\ref{fig_model}. It consists of three stages of tuning curriculum learning. Curriculum learning \cite{wang2021survey} breaks down complex tasks into simpler ones and trains models in an increasing order of difficulty. This approach mimics the way humans learn by first mastering basic concepts before progressing to more complex knowledge.

From the previous theoretical analysis, we find that the three types of graphs have a progressive relationship. In a KG, entities and relations are represented as triples, which can be understood as event nodes in an EKG to some extent. EKG further explores the relationships between event nodes, while a CKG can be seen as a generalization of EKG, based on more universal commonsense knowledge.

Therefore, the tuning framework is divided into three stages following a curriculum learning approach: the \textit{KG empowerment stage}, the \textit{EKG enhancement stage}, and the \textit{CKG generalization stage}. After the KG empowerment stage, we obtain the G-Micro model, which is expected to handle basic sub-tasks related to KG, such as  handling various entity and relation extraction tasks. However, GKG nodes and relationships may include dynamic knowledge. Next, in the EKG enhancement stage, we utilize EKG-related sub-tasks datasets to further empower GKG-LLM on the basis of G-Micro, resulting in the G-Mid model, capable of handling sub-tasks involving dynamic knowledge. Furthermore, in the CKG generalization stage, we inject CKG-related sub-tasks and counter task data into the G-Mid model, generalizing the task handling capability of KG to broader scenarios, ultimately resulting in the GKG-LLM model.

\paragraph{KG empowerment stage}

At this stage, we only inject the KG sub-task dataset into LLMs, and the training loss function is defined as cross-entropy loss:

\begin{equation}
\mathcal{L}_{\text{CE}}=-\sum\limits_{i}p\left(y_i\right)\log p_{\theta}\left(\hat{y_i} \mid s_i ; x_i\right ),
\end{equation}

\noindent where \( p_{\theta}\) represents the tunable LLM with parameters \(\theta\), initialized from the base model. The instruction \( s_i \) is concatenated with the input \(x_i\) denotes the prompt format to LLMs. \(\hat{y_i}\) is the predicted output, while \(y_i\) represents the ground truth.

\paragraph{EKG Enhancement Stage}

At this stage, we inject knowledge about dynamic nodes and relationships to enhance the model’s capability. Specifically, we train the G-Micro model from the first stage using the EKG sub-task dataset. This process expands the model’s understanding of complex graphs, enabling it to handle dynamic nodes and relationships with temporal dependencies and causal features, improving its adaptability to changing data and laying a foundation for the subsequent stages. The loss function is the same as in the first stage.

\paragraph{CKG Generalization Stage}

Real-world scenarios go beyond static knowledge and specific events, encompassing commonsense knowledge for a broader understanding. Therefore, at this stage, we train the G-Mid model from the second stage using the CKG sub-task dataset to enhance its generalization and applicability. This expands the model's commonsense knowledge, enabling it to excel in open-ended and complex reasoning tasks~\cite{xu2025large}. The model becomes more practical and effective in real-world scenarios, ultimately resulting in the GKG-LLM.

This study conducts extensive testing and analysis on three types of data: In-domain, OOD and counter task data. Detailed implementation specifics is discussed in the following sections.

\subsection{Data Collection and Preparation} \label{2.2}

As a comprehensive dataset encompassing the GKG construction tasks, it requires extensive datasets for each sub-task across the three types of graphs. Additionally, it is necessary to perform reasonable partitioning of the various datasets and format them to prepare for the unified GKG construction framework.

The overview of data distribution of all of GKG sub-tasks is shown as Figure~\ref{data_distribution}. The GKG dataset is \( \mathcal{D}_G = \mathcal{D}_{KG}\bigcup \mathcal{D}_{EKG} \bigcup \mathcal{D}_{CKG} \bigcup \mathcal{D}_{ct}\). Here, \(\mathcal{D}_{KG}\) includes the sub-tasks of KG  such as relation extraction and entity-relation joint extraction; For \(\mathcal{D}_{EKG}\), sub-tasks include sentence-level event detection, document-level event argument extraction, and event temporal relation extraction; And for \(\mathcal{D}_{CKG}\), sub-tasks include summary generation and text inference. \(\mathcal{D}_{ct}\) refers to a structure-to-text dataset, specifically the WebNLG task and dataset used for natural language generation, designed to serve as a counter-task for all GKG sub-tasks to prevent overfitting and enhance generalization without compromising the primary performance. Finally,  we obtain \( \mathcal{D}_G\) of $\sim$806K pieces for training and $\sim$140K pieces for testing. Details of each dataset are attached in Appendix~\ref{DataCollection}. The details of each sub-task are provided in Appendix~\ref{sub-tasks}.

After data collection, we format each piece \( i\) of the GKG dataset into a unified format, which includes \(ID\), instruction \( s_i \), few-shot \( fs\) / zero-shot \( zs\) , input \(x_i\), and output \(y_i\). Details of the data format and few-shot organization can be found in Appendix~\ref{DataFormat}.

\subsection{Training Strategy} \label{2.3}

To effectively fine-tune our model on the unified dataset, we employ the LoRA+ \cite{hayou2024lora+} technique, an advanced version of Low-Rank Adaptation (LoRA), which has shown great promise in parameter-efficient fine-tuning (PEFT). LoRA+ adapts only a small subset of model parameters, reducing computational costs while maintaining high performance. By leveraging low-rank matrix approximations, LoRA+ allows us to efficiently update the model parameters without the need for extensive computational resources. Formally, LoRA+ modifies the weight matrix \( W \) in the neural network as follows:

\begin{table*}[t]
\centering
\footnotesize
\resizebox{\linewidth}{!}{
\begin{tabular}{lcc|ccc|ccccc}
    \toprule
    \multirow{2}{*}{\textbf{Graphs}} & \multirow{2}{*}{\textbf{Tasks}} & \multirow{2}{*}{\textbf{Datasets}} & GPT- & Claude- & Gemini-& \multicolumn{2}{c}{LlaMA-} & Single- & Integrated- & GKG- \\
    & & & 4 & 3 & 1.5 & 2-GKG & 3-Instruct & SFT & SFT & LLM \\
    \midrule
    \multirow{6}{*}{\textbf{KG}} & SRE & NYT & 64.94 & 66.76 & 68.59 & 78.18 & 55.12 & 74.39 & \underline{79.32} & \textbf{80.63} \\
    & \multirow{2}{*}{FRE} & FewRel & 26.28 & 27.45 & 30.20 & \underline{89.45} & 22.64 & 78.65 & 86.74 & \textbf{90.48} \\
    & & TACRED & 18.85 & 20.23 & 22.43 & \underline{86.71} & 12.74 & 70.66 & 84.66 & \textbf{88.96} \\
    & DRE & DOCRED & 38.84 & 36.28 & 42.63 & 83.18 & 34.63 & 74.53 & \underline{83.61} & \textbf{85.71} \\
    & \multirow{2}{*}{JE\&RE} & FewRel & 6.32 & 5.44 & 7.52 & \textbf{42.05} & 3.20 & 26.76 & 30.56 & \underline{34.32} \\
    & & NYT & 6.22 & 5.85 & 8.36 & \textbf{53.33} & 0.0 & 40.16 & 48.66 & \underline{52.27}\\
    \midrule
    \multirow{15}{*}{\textbf{EKG}} & SED & ACE2005 & 17.50 & 8.57 & 22.40 & 32.47 & 0.0 & 22.74 & \underline{34.32} & \textbf{80.63} \\
    & DED & WIKIEVENTS & 16.54 & 9.14 & 14.87 & 24.87 & 18.62 & \underline{29.59} & 23.84 & \textbf{39.86} \\
    & \multirow{2}{*}{DEAE} & WIKIEVENTS & 42.58 & 53.41 & 47.69 & \underline{70.46} & 41.76 & 63.38 & 69.30 & \textbf{75.22} \\
    &  & RAMS & 13.84 & 5.70 & 38.49 & 48.33 & 30.74 & \underline{53.43} & 52.09 & \textbf{63.62} \\
    & \multirow{6}{*}{ETRE} & MATRES & 39.97 & 36.62 & 38.51 & \underline{62.94} & 22.79 & 37.91 & 44.26 & \textbf{71.51} \\
    & & ESL & 64.24 & 47.65 & 42.18 & 68.96 & 21.67 & \underline{74.06} & 67.63 & \textbf{75.33} \\
    & & TB-Dense & 43.73 & 36.58 & 42.43 & \underline{52.89} & 36.55 & 49.30 & 51.23 & \textbf{53.54} \\
    & & Causal-TB & 6.67 & 8.01 & 8.74 & 42.79 & 16.43 & 37.35 & \textbf{49.83} & \underline{45.26} \\
    & & MAVEN-ERE & 43.80 & 21.73 & 42.10 & 71.55 & 40.29 & 37.35 & \underline{75.44} & \textbf{81.95} \\
    & & TCR\(^*\) & 15.43 & 18.74 & \underline{25.34} & 24.88 & 24.71 & 20.68 & 22.09 & \textbf{26.45} \\
    & \multirow{3}{*}{ECRE} & ESL & 28.57 & 19.26 & 55.21 & 75.33 & 26.33 & 62.92 & \underline{78.74} & \textbf{84.89} \\
    & & MAVEN-ERE & 51.98 & 11.36 & 43.38 & 76.48 & 13.37 & 78.91 & \underline{88.59} & \textbf{90.18} \\
    & & Causal-TB\(^*\) & 39.67 & 41.23 & 43.44 & 33.94 & 30.02 & 48.41 & \underline{48.80} & \textbf{55.79} \\
    & \multirow{2}{*}{ESRE} & HiEve & 38.81 & 30.92 & 48.83 & 55.60 & 48.61 & 57.64 & \underline{58.01} & \textbf{58.61}\\
    & & MAVEN-ERE & 40.09 & 13.12 & 38.09 & \underline{44.37} & 33.49 & 39.11 & 37.30 & \textbf{48.49} \\
    \midrule
    \multirow{7}{*}{\textbf{CKG}} & NER & CoNLL & 15.94 & 14.46 & 18.27 & \underline{77.50} & 15.60 & 64.74 & 70.53 & \textbf{82.30} \\
    & \multirow{2}{*}{AG}\(\dagger\) & CNNDM & 30 & 28 & 22 & \underline{36} & 18 & 35 & 35 & \textbf{45} \\
    & & XSum & \underline{33} & 26 & 29 & 28 & 9 & 24 & 30 & \textbf{38} \\
    & \multirow{2}{*}{LI} & SNLI & 51.26 & 47.56 & 60.38 & 69.51 & 44.50 & 87.09 & \textbf{89.35} & \underline{89.03} \\
    & & MNLI & 81.80 & 39.33 & 48.80 & 58.97 & 53.70 & \textbf{86.78} & 84.62 & \underline{86.35} \\
    & \multirow{2}{*}{TC} & R8\(^*\) & \textbf{72.26} & 36.43 & 66.58 & 65.27 & 58.89 & 28.83 & 58.64 & \underline{69.33} \\
    & & R52 & 82.18 & 83.75 & 80.63 & \textbf{94.16} & 29.68 & 89.02 & 88.81 & \underline{90.34} \\
    \midrule
    \textit{Counter} & NLG\(\dagger\) & WebNLG & 78 & 65 & 76 & \underline{83} & 15 & 80 & 80 & \textbf{85} \\
    \midrule
    \rowcolor{lightgray}
    \multicolumn{3}{c|}{\textbf{Average Performance}}  & 38.25 & 29.81 & 39.07 & 59.70 & 26.83 & 52.97 & \underline{60.41} & \textbf{67.90} \\
    \bottomrule
\end{tabular}}
\caption{Performance comparison across various datasets and tasks. The best result for each sub-task is highlighted in bold, while the second-best result is underlined. The OOD datasets are starred by *. \(\dagger\) means the task is evaluated by metric Rough-L of percentage. The results for GPT-4, Claude-3, and Gemini-1.5 are obtained via their respective APIs. LlaMA-2-GKG, LlaMA-3-Instruct, Single-SFT, and Integrated-SFT are implemented by us. The GKG-LLM column represents the final model obtained after three-stage tuning.}
\label{mainResults}
\end{table*}

\begin{equation}
W' = W + \Delta W,
\end{equation}

\noindent where \( \Delta W = AB \), with \( A \in \mathbb{R}^{d \times r} \) and \( B \in \mathbb{R}^{r \times k} \). Here, \( d \) is the dimension of the input, \( k \) is the dimension of the output, and \( r \) is the rank of the adaptation matrices, which is much smaller than both \( d \) and \( k \), making the adaptation parameter-efficient.
To make better use of limited resources for training the model, the advancement of LoRA+ is reflected, as shown in Equation~\ref{equ3}, in the use of different update hyperparameters \(\eta_A\) and \(\eta_B\) for the two low-rank matrices \( A \) and \( B \):

\begin{equation}
\left\{
\begin{aligned}
    & A = A - \eta_A G_A \\
    & B = B - \eta_B G_B.
\end{aligned}
\right.
\label{equ3}
\end{equation}

\noindent This approach accelerates convergence and effectively demonstrates the efficient and adaptive capabilities of GKG-LLM in handling GKG construction sub-tasks.

In summary, our training process harnesses the strengths of LoRA+ for efficient fine-tuning while experimenting with diverse data utilization strategies to optimize model performance for comprehensive GKG construction. This approach ensures that our model not only learns effectively from the data but also adapts seamlessly to various NLP tasks within GKG.

\section{Experiments}

In this section, we thoroughly evaluate the performance of GKG-LLM across three data settings, including in-sample data, counter-task data, and out-of-distribution data. The baseline methods and evaluation metrics are presented in Section~\ref{3.1}, while the main experimental results are presented in Sections~\ref{3.2}. The stage generalization results are presented in Appendix~\ref{3.3}. Hyper-parameter settings are provided in Appendix~\ref{stages}.

\subsection{Baselines and Metrics} \label{3.1}

To perform a comprehensive evaluation, the final version of GKG-LLM is compared with two main categories of existing baselines: close-source baselines and open-source baselines.

For closed-source baselines, we access the model through the OpenAI API, specifically using the gpt-4-turbo-preview version\footnote{https://openai.com/api/}, and the Anthropic API to access the Claude-3-Opus version\footnote{https://www.anthropic.com/api} for evaluation. We also use the Google API to access the Gemini-1.5-Pro version\footnote{https://deepmind.google/technologies/gemini/pro/} for evaluation.

For open-source baselines, we conduct experiments on two foundations: LlaMA-2-Chat\footnote{https://huggingface.co/meta-llama/Llama-2-7b-chat-hf} and LlaMA-3-Instruct\footnote{https://huggingface.co/meta-llama/Meta-Llama-3-8B-Instruct}. The LlaMA-2-GKG is fine-tuned from Llama-2-Chat, while LlaMA-3-Instruct serves as the foundation for GKG-LLM and also acts as a baseline. This model is fine-tuned to fit a specific graph, serving as a strong baseline. Our integrated SFT method trains all datasets from the three types of graphs simultaneously.

Referencing the general evaluation metrics for each sub-task, for abstraction generation and structure-to-text tasks, the Rough-L metric is used, while all other tasks employ the F1 score as the evaluation metric.

\subsection{Main Results} \label{3.2}

In this section, we thoroughly evaluate the performance of GKG-LLM on in-domain, OOD, and counter tasks. Specifically, as detailed in Table~\ref{mainResults}, we assess its performance across various sub-tasks in the three types of graphs. Compared to the baseline, the results demonstrate the effectiveness and practicality of GKG-LLM on the construction of all three graph types across in-domain, OOD, and counter-task data.

\paragraph{KG Sub-task Datasets}

KG sub-task datasets focus on various types of relation extraction, including sentence-level relation extraction, few-shot relation extraction, and entity relation extraction, etc. Compared to the three closed-source LLMs, GKG-LLM achieves the best performance, with a minimum performance improvement of 12.04\%. Additionally, when compared to a model tuned solely with KG sub-task datasets, GKG-LLM demonstrates a minimum performance gain of 7.6\%. Across all baselines, GKG-LLM consistently achieves either the best or the second-best performance.

\paragraph{EKG Sub-task Datasets}

EKG sub-task datasets primarily include event detection, event argument extraction, and event relation extraction. Compared to the three closed-source LLMs, GKG-LLM achieves the best performance, with a minimum improvement of 9.88\%. An interesting observation is that the Integrated SFT model achieves the second-best performance in half of the tasks; however, GKG-LLM still consistently performs either the best or the second-best overall. Another interesting point is that in the OOD datasets, specifically the TCR dataset for the ETRE sub-task and the Causal-TB dataset for the ECRE sub-task, GKG-LLM outperforms the second-best baseline by 1.11\% and 6.99\%, respectively, demonstrating its strong generalization capability on OOD data.

\paragraph{CKG Sub-task Datasets}

For the CKG sub-task dataset, the focus is closer to common-sense nodes and relations reasoning, involving tasks such as abstract generation and language inference. For the R8 dataset in the Text Classification sub-task, which serves as an OOD dataset, GPT-4 achieves the best performance, attributed to its exceptional capabilities in language understanding. Even so, GKG-LLM still achieves the second-best performance. Since CKG closely resembles real-world commonsense scenarios, both LlaMA-2-GKG and Single-SFT also demonstrates strong results. However, overall, GKG-LLM consistently maintains either the best or the second-best performance.

GKG-LLM achieves the best performance on the WebNLG dataset for the Natural Language Generation (NLG) task, surpassing the strongest baseline by 2\%, further highlighting its strong structure-to-text capabilities. It consistently performs at the best or second-best level across all GKG sub-tasks, with an average improvement of 7.49\% over the strongest baseline. Additionally, its strong performance on OOD data demonstrates its ability to generalize effectively to unseen data distributions, with ablation studies and OOD analysis detailed in Section~\ref{anlysis}.

\subsection{Exploration of Three Stages} \label{}

As discussed in Section~\ref{1}, a triple in a KG can, to some extent, be considered as a node in an EKG, while the triples in EKG and CKG are linked through the relationship between the concrete and the abstract. Theoretically, there exists a progressive relationship among these three types of graphs, which serves as the theoretical basis for our three-stage fine-tuning framework. Therefore, this subsection will explore the performance of the three types of graphs under different fine-tuning sequences, as well as the performance of the intermediate versions of our three-stage fine-tuning framework on the sub-tasks of the three types of graphs.

\begin{figure}[h]
    \centering
    \includegraphics[width=\columnwidth]{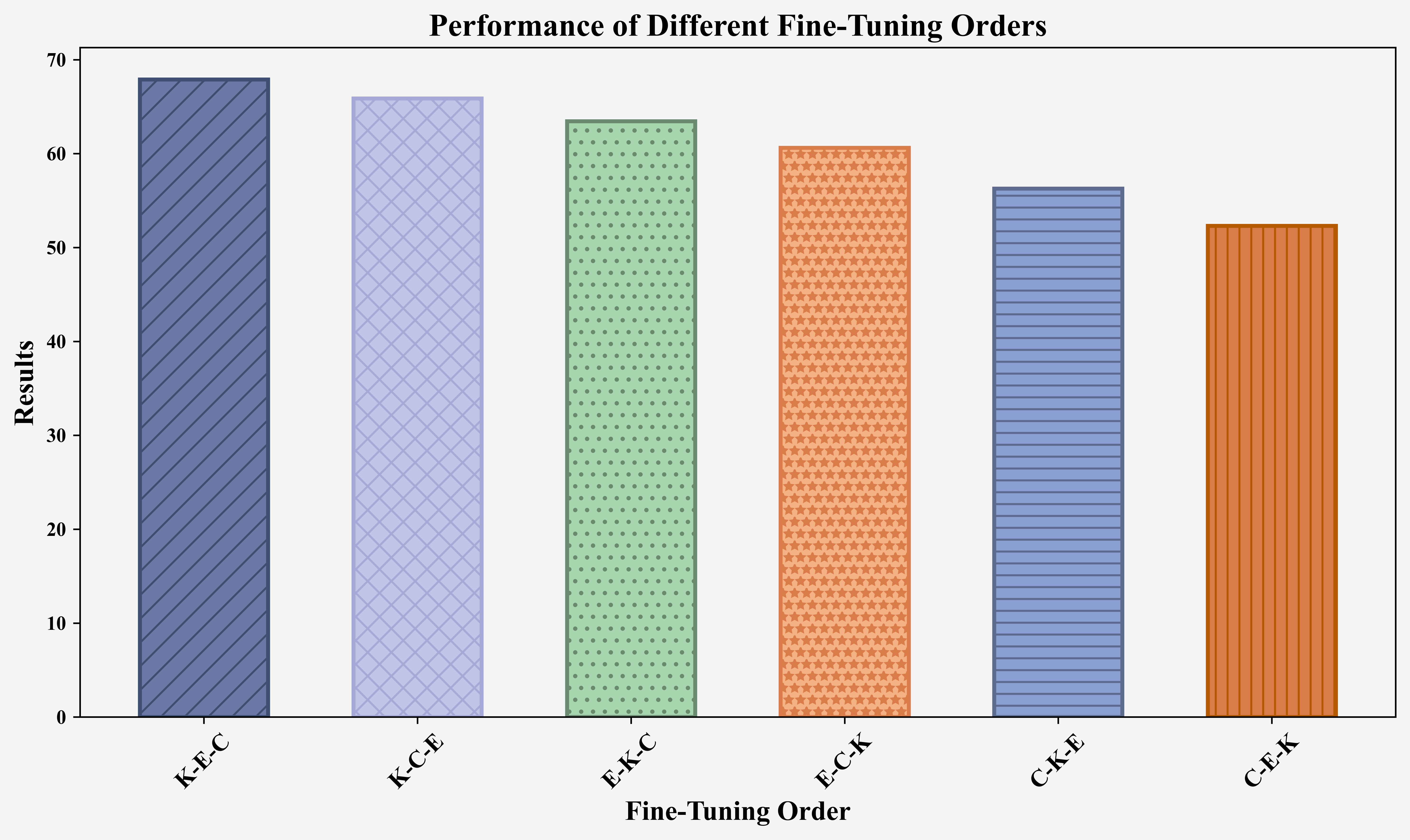}
    \caption{Results of different fine-tuning orders. ``K-E-C" means the fine-tuning order is KG, EKG and CKG. The following sets of experiments are similar to this one.}
    \label{orders}
\end{figure}

As shown in Figure~\ref{orders}, the three types of graphs show varying performance in terms of average performance across all tasks under different fine-tuning sequences. The ``K-E-C" sequence adopted in this study demonstrates the best performance, further confirming the theoretical correctness and experimental effectiveness of our three-stage fine-tuning sequence.

\begin{figure}[h]
    \centering
    \includegraphics[width=\columnwidth]{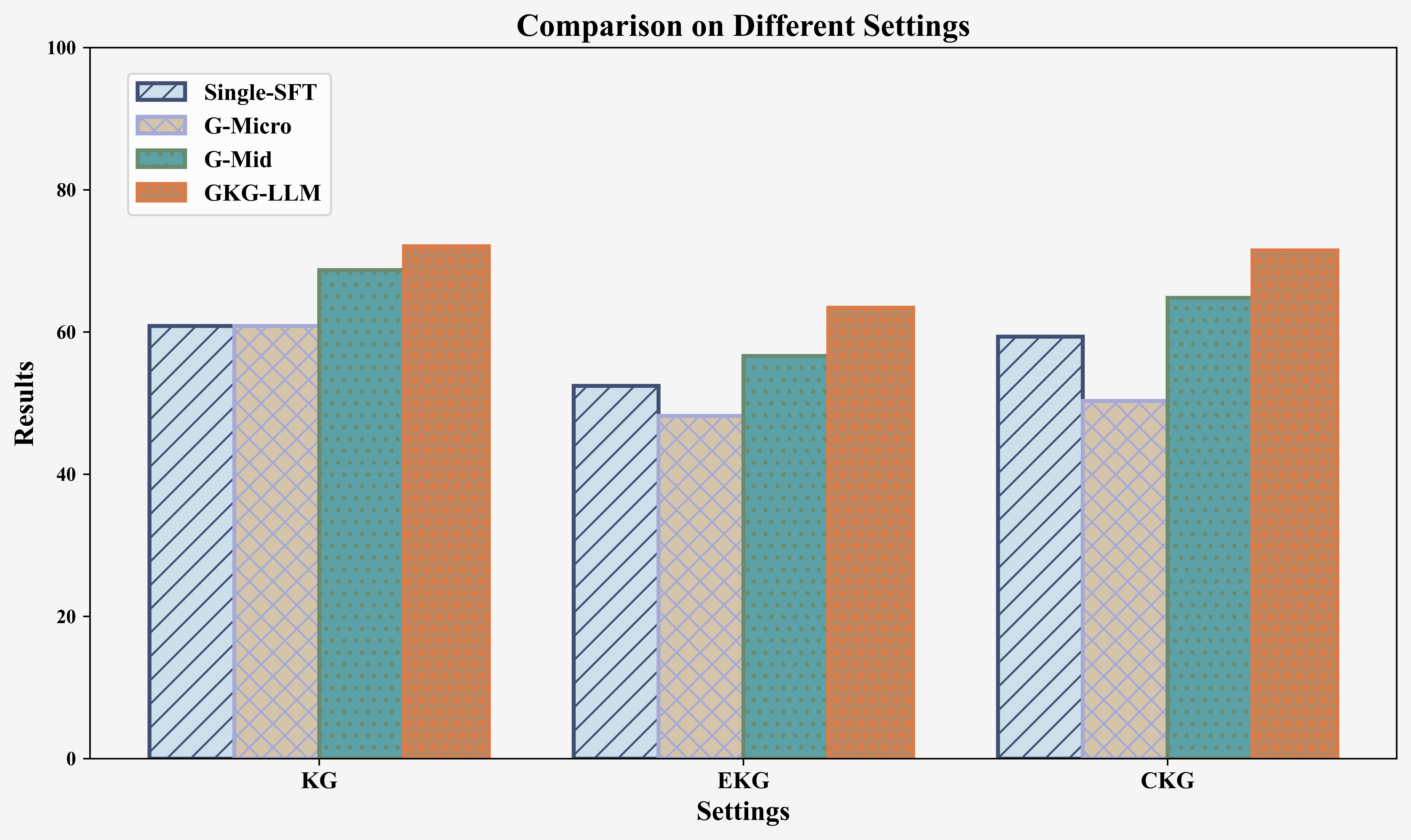}
    \caption{Fine-tuning with a single type of graph and performance of different intermediate version in the GKG-LLM.}
    \label{intermediate}
\end{figure}

Figure~\ref{intermediate} presents the performance of the single SFT model and the three-stage models across the KG, EKG, and CKG sub-tasks. In each sub-task, the results improve as the fine-tuning progresses through the three stages. Compared to single-SFT, our GKG-LLM framework demonstrates better performance, validating the practicality of the three-stage fine-tuning approach.

\section{Analysis} \label{anlysis}

In this section, we introduce the ablation study in Section~\ref{4.1} and provide a comprehensive analysis and explanation of the OOD data in Section~\ref{4.2}. An analysis of data scaling in training is introduced in Section~\ref{4.3}. The evaluation of the optimal model under various hyper-parameter settings is presented in Appendix~\ref{hyperparameters}.

\begin{table}[htbp]
    \centering
    \small 
    \begin{tabular}{l|cccc}
        \toprule
        \textbf{Variation} & \textbf{KG} &  \textbf{EKG} & \textbf{CKG} &  \textbf{Avg.} \\
        \midrule
        $\mathcal{P}$ & 72.06 & 63.42 & 71.48 & 67.90\\
        \midrule
        $\mathcal{P}_{\text{si}}$ & 68.46 & 59.34 & 69.10 & 64.33 \\
        $\Delta$ & \textcolor{deepgreen}{(-3.60)} & \textcolor{deepgreen}{(-4.08)} & \textcolor{deepgreen}{(-2.38)} & \textcolor{deepgreen}{(-3.57)} \\
        $\mathcal{P}_{\text{zs}}$ & 65.17 & 55.09 & 66.05 & 60.06 \\
        $\Delta$ & \textcolor{deepgreen}{(-6.89)} & \textcolor{deepgreen}{(-8.33)} & \textcolor{deepgreen}{(-5.43)} & \textcolor{deepgreen}{(-7.84)} \\
        $\mathcal{P}_{\text{si+zs}}$ & 62.44 & 52.26 & 64.66 & 58.15 \\
        $\Delta$ & \textcolor{deepgreen}{(-9.62)} & \textcolor{deepgreen}{(-11.16)} & \textcolor{deepgreen}{(-6.82)} & \textcolor{deepgreen}{(-9.75)} \\
        \bottomrule
    \end{tabular}
    \caption{Performance comparison of different prompt strategies on the evaluation metrics. $\mathcal{P}$ denotes full prompts, $\mathcal{P}_{\text{si}}$ refers to a single instruction regardless of diversity, $\mathcal{P}_{\text{zs}}$ represents zero-shot only, and $\mathcal{P}_{\text{si+zs}}$ combines single instruction with zero-shot prompting.}
    \label{tab:prompt_content_effectiveness}
\end{table}

\subsection{Ablation Studies} \label{4.1}

In this section, we present the ablation study for three different prompt strategies: (1) using only a single instruction to construct the prompt format, (2) using only zero-shot prompts without employing any few-shot examples, and (3) removing both strategies simultaneously. We compare the performance across three types of graphs and the overall dataset, with the comparison results shown in Table~\ref{tab:prompt_content_effectiveness}.
Examples of different types of prompts can be found in the respective sections of Appendix~\ref{DataFormat}.

The results show that removing the diversity of instructions causes a noticeable performance drop, as diverse instructions better reflect real-world scenarios where different questioners have unique styles, requiring the model to adapt to various instruction formats. Removing the few-shot learning strategy lead to an even greater performance degradation, as LLMs lost their ability to perform in-context learning and relies only on inherent capabilities, affecting their ability to generate the corresponding elements or relationships. The most  performance drop occurs when both strategies are removed, highlighting that the advantages of these strategies are cumulative, further validating the superiority and effectiveness of our data construction strategy.

\subsection{OOD Analysis} \label{4.2}

This section specifically discusses the performance of GKG-LLM on OOD datasets. As introduced in Section~\ref{2.1}, our data is divided into three parts, with the OOD portion deliberately excluded during the initial training design, meaning that GKG-LLM has never encountered these types of data before. Therefore, the performance on this part serves as an indicator of our model's generalization ability from the perspective of OOD data.

\begin{figure}[t]
    \includegraphics[width=\columnwidth]{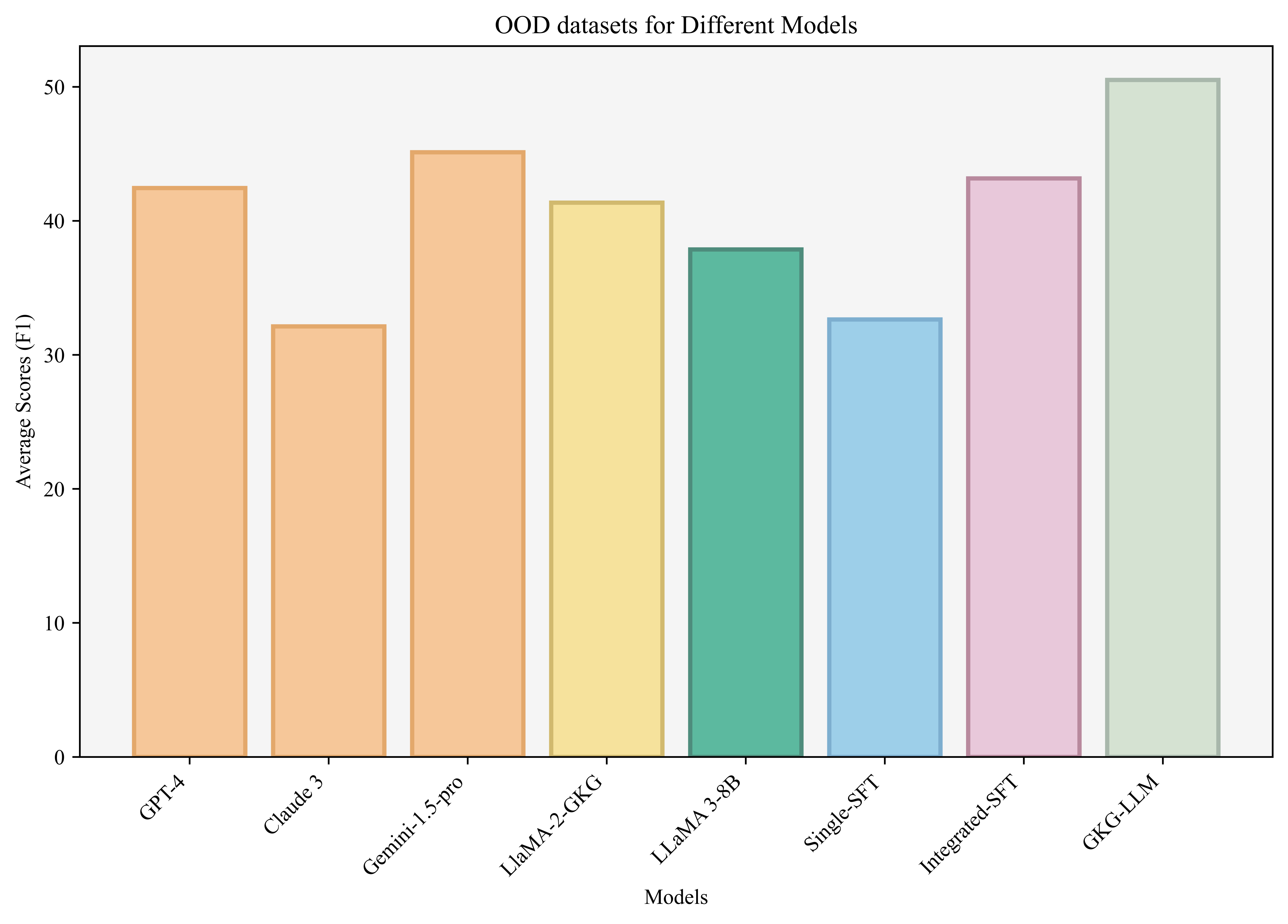}
    \caption{The average performance on OOD datasets, consisting TCR, Causal-TB and R8 datasets.}
    \label{data_OOD}
\end{figure}

As shown in Figure~\ref{data_OOD}, overall, our method achieves the best performance, reaching 50.52\%, which is 5.40\% higher than the second-best model, Gemini-1.5-pro. Despite the fact that these data points were entirely unfamiliar to both closed-source LLMs and our tuned open-source LLMs, our model still demonstrates strong robustness and effectiveness.

\subsection{Analysis on Different Data Scaling} \label{4.3}

This section explores the impact of different data scales on model performance. The model is trained using 10\%, 20\%, 40\%, 60\%, 80\%, and 100\% of the data, sampled from the three types of graph sub-tasks separately. The results show that as the data proportion increases, model performance improves progressively, with performance being limited at 10\%, improving at 20\% and 40\%, and continuing to enhance at 60\% and 80\%, reaching near-optimal performance at 100\%.

\begin{figure}[h]
    \centering
    \includegraphics[width=\columnwidth]{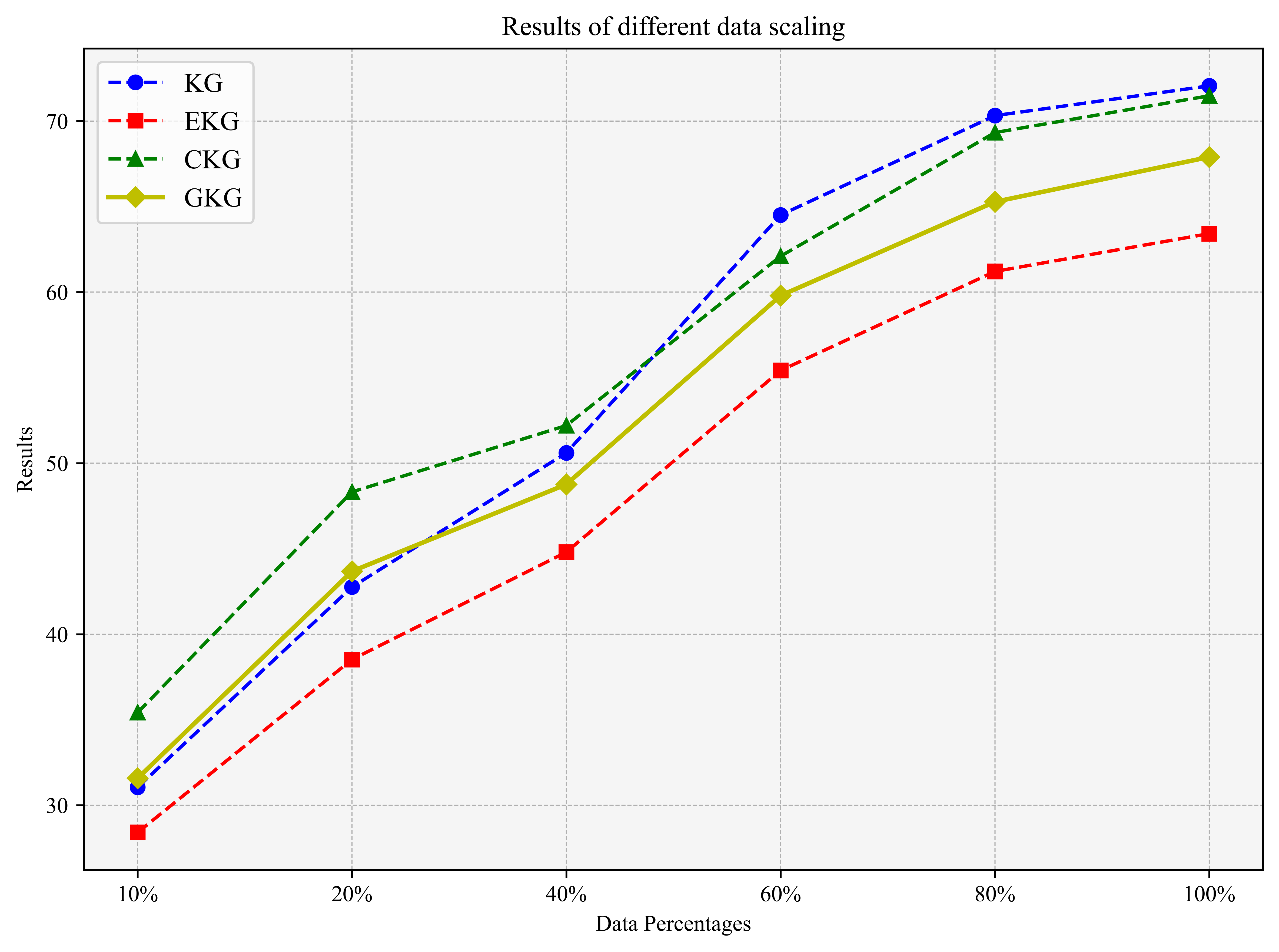}
    \caption{Results of training with different proportions of complete data.}
    \label{fig:data_percentage_results}
\end{figure}

Figure \ref{fig:data_percentage_results} shows that as the data volume increases, the model’s average scores across all tasks gradually improve. Notably, the average scores for the three types of graph sub-tasks follow similar trends, with diminishing performance gains beyond 80\% data usage, indicating a saturation point where the additional data brings marginal benefits.

\section{Related Works}

This section introduces two types of related work. Section~\ref{5.1} covers three typical tasks within GKG sub-tasks, while Section~\ref{5.2} discusses research related to LLMs.

\subsection{GKG Sub-tasks} \label{5.1}

In this section, we introduce a representative task for each of the three types of graphs: the entity-relation joint extraction task in the KGs, the document-level event argument extraction task in the EKGs, and the abstract generation task in the CKGs.

\textbf{Entity-relation joint extraction} task has been a focus in the domain of knowledge graph construction, as it aims to simultaneously extract entities and their relationships from unstructured text. Current state-of-the-art methods leverage transformer architecture to model interactions between entities within sentences or documents, which provides further performance gains \cite{sui2023joint}. 
\textbf{Document-level event argument extraction} aims to extract the arguments of events from long texts to better understand complex event relations and event chains. Pre-trained models such as BERT have been widely employed in event extraction tasks. By combining pre-trained knowledge with task-specific fine-tuning, these models have proven effective in understanding complex contexts \cite{zhang2024semantic}.
\textbf{Abstract generation} particularly with the rise of pre-trained transformer-based models. A recent state-of-the-art approach by \cite{gao2023comparing} utilizes a combination of pre-trained language models and reinforcement learning to enhance the quality of generated abstracts.

\subsection{Large Language Models} \label{5.2}

With the emergence of closed-source and open-source LLMs represented by GPT4 \cite{achiam2023gpt} and LlaMA-3 \cite{dubey2024llama}, respectively, a large amount of research has focused on these models. This section introduces some of the work based on close-source and open-source LLMs.

Research based on closed-source LLMs typically involves evaluating these large models \cite{gandhi2024understanding} and integrating them with traditional tasks. For example, such studies may focus on enhancing certain aspects of conventional natural language tasks \cite{zheng2023can} or providing new perspectives for text analysis \cite{savelka2023can}.
The study by \cite{xu2023symbol} using LlaMA-2 as the foundation, explores the possibility of a unified approach to symbol-centric tasks through full fine-tuning and extend this approach to generalize to natural language-centric tasks. A survey by \cite{zhang2023instruction} introduce various paradigms of instruction fine-tuning for LLMs, providing a comprehensive overview of its advantages, limitations, and implementation methods.

However, up to now, no study has integrated the broad task of GKG construction. This research unifies such tasks from both the task and data perspectives by fine-tuning open-source LLMs.

\section{Conclusion}
\label{sec:bibtex}

This study proposes a new task for building GKG. It represents the first collection approached from the unified perspective in terms of data, and the first unified construction of three types of graphs from the task perspective. This task addresses two issues: obstacles arising from differences between tasks, and the neglect of intrinsic connections among different types of graphs. To address these challenges, we propose a three-stage curriculum learning framework that iteratively injects sub-task knowledge from KG, EKG, and CKG into GKG-LLM, aiming for broad and outstanding performance in GKG construction. Extensive experiments demonstrate the effectiveness and robustness of the GKG-LLM approach. The models and data from this study will be fully released upon acceptance of the paper.
In the future, we will expand the application of GKG-LLM into a broader range of scenarios, such as intelligent healthcare~\cite{DBLP:journals/inffus/HeMLRLFC25,DBLP:journals/inffus/LinZMHMHPCF25}, to enhance its utility and impact.


\bibliographystyle{named}
\bibliography{ijcai25}

\begin{thebibliography}{}

\bibitem[\protect\citeauthoryear{Achiam \bgroup \em et al.\egroup }{2023}]{achiam2023gpt}
Josh Achiam, Steven Adler, Sandhini Agarwal, Lama Ahmad, Ilge Akkaya, Florencia~Leoni Aleman, Diogo Almeida, Janko Altenschmidt, Sam Altman, Shyamal Anadkat, et~al.
\newblock Gpt-4 technical report.
\newblock {\em arXiv preprint arXiv:2303.08774}, 2023.

\bibitem[\protect\citeauthoryear{Alt \bgroup \em et al.\egroup }{2020}]{alt2020tacred}
Christoph Alt, Aleksandra Gabryszak, and Leonhard Hennig.
\newblock Tacred revisited: A thorough evaluation of the tacred relation extraction task.
\newblock {\em arXiv preprint arXiv:2004.14855}, 2020.

\bibitem[\protect\citeauthoryear{Camburu \bgroup \em et al.\egroup }{2018}]{camburu2018snli}
Oana-Maria Camburu, Tim Rockt{\"a}schel, Thomas Lukasiewicz, and Phil Blunsom.
\newblock e-snli: Natural language inference with natural language explanations.
\newblock {\em Advances in Neural Information Processing Systems}, 31, 2018.

\bibitem[\protect\citeauthoryear{Chan \bgroup \em et al.\egroup }{2024}]{chan2024exploring}
Chunkit Chan, Cheng Jiayang, Weiqi Wang, Yuxin Jiang, Tianqing Fang, Xin Liu, and Yangqiu Song.
\newblock Exploring the potential of chatgpt on sentence level relations: A focus on temporal, causal, and discourse relations.
\newblock In {\em Findings of the Association for Computational Linguistics: EACL 2024}, pages 684--721, 2024.

\bibitem[\protect\citeauthoryear{Chen \bgroup \em et al.\egroup }{2021}]{chen2021dialogsum}
Yulong Chen, Yang Liu, Liang Chen, and Yue Zhang.
\newblock Dialogsum: A real-life scenario dialogue summarization dataset.
\newblock {\em arXiv preprint arXiv:2105.06762}, 2021.

\bibitem[\protect\citeauthoryear{Dubey \bgroup \em et al.\egroup }{2024}]{dubey2024llama}
Abhimanyu Dubey, Abhinav Jauhri, Abhinav Pandey, Abhishek Kadian, Ahmad Al-Dahle, Aiesha Letman, Akhil Mathur, Alan Schelten, Amy Yang, Angela Fan, et~al.
\newblock The llama 3 herd of models.
\newblock {\em arXiv preprint arXiv:2407.21783}, 2024.

\bibitem[\protect\citeauthoryear{Ebner \bgroup \em et al.\egroup }{2020}]{ebner2020multi}
Seth Ebner, Patrick Xia, Ryan Culkin, Kyle Rawlins, and Benjamin Van~Durme.
\newblock Multi-sentence argument linking.
\newblock In {\em Proceedings of the 58th Annual Meeting of the Association for Computational Linguistics}, pages 8057--8077, 2020.

\bibitem[\protect\citeauthoryear{Gandhi \bgroup \em et al.\egroup }{2024}]{gandhi2024understanding}
Kanishk Gandhi, Jan-Philipp Fr{\"a}nken, Tobias Gerstenberg, and Noah Goodman.
\newblock Understanding social reasoning in language models with language models.
\newblock {\em Advances in Neural Information Processing Systems}, 36, 2024.

\bibitem[\protect\citeauthoryear{Gao \bgroup \em et al.\egroup }{2023}]{gao2023comparing}
Catherine~A Gao, Frederick~M Howard, Nikolay~S Markov, Emma~C Dyer, Siddhi Ramesh, Yuan Luo, and Alexander~T Pearson.
\newblock Comparing scientific abstracts generated by chatgpt to real abstracts with detectors and blinded human reviewers.
\newblock {\em NPJ Digital Medicine}, 6(1):75, 2023.

\bibitem[\protect\citeauthoryear{Gardent \bgroup \em et al.\egroup }{2017}]{gardent2017webnlg}
Claire Gardent, Anastasia Shimorina, Shashi Narayan, and Laura Perez-Beltrachini.
\newblock The webnlg challenge: Generating text from rdf data.
\newblock In {\em 10th International Conference on Natural Language Generation}, pages 124--133. ACL Anthology, 2017.

\bibitem[\protect\citeauthoryear{Ge and Moh}{2017}]{ge2017improving}
Lihao Ge and Teng-Sheng Moh.
\newblock Improving text classification with word embedding.
\newblock In {\em 2017 IEEE International Conference on Big Data (Big Data)}, pages 1796--1805. IEEE, 2017.

\bibitem[\protect\citeauthoryear{Glava{\v{s}} \bgroup \em et al.\egroup }{2014}]{glavavs2014hieve}
Goran Glava{\v{s}}, Jan {\v{S}}najder, Parisa Kordjamshidi, and Marie-Francine Moens.
\newblock Hieve: A corpus for extracting event hierarchies from news stories.
\newblock 2014.

\bibitem[\protect\citeauthoryear{Grishman \bgroup \em et al.\egroup }{2005}]{grishman2005nyu}
Ralph Grishman, David Westbrook, and Adam Meyers.
\newblock Nyu’s english ace 2005 system description.
\newblock {\em Ace}, 5(2), 2005.

\bibitem[\protect\citeauthoryear{Gubelmann \bgroup \em et al.\egroup }{2024}]{gubelmann2024capturing}
Reto Gubelmann, Ioannis Katis, Christina Niklaus, and Siegfried Handschuh.
\newblock Capturing the varieties of natural language inference: A systematic survey of existing datasets and two novel benchmarks.
\newblock {\em Journal of Logic, Language and Information}, 33(1):21--48, 2024.

\bibitem[\protect\citeauthoryear{Han \bgroup \em et al.\egroup }{2018}]{han2018fewrel}
Xu~Han, Hao Zhu, Pengfei Yu, Ziyun Wang, Yuan Yao, Zhiyuan Liu, and Maosong Sun.
\newblock Fewrel: A large-scale supervised few-shot relation classification dataset with state-of-the-art evaluation.
\newblock {\em arXiv preprint arXiv:1810.10147}, 2018.

\bibitem[\protect\citeauthoryear{Han \bgroup \em et al.\egroup }{2019}]{han2019deep}
Rujun Han, I~Hsu, Mu~Yang, Aram Galstyan, Ralph Weischedel, Nanyun Peng, et~al.
\newblock Deep structured neural network for event temporal relation extraction.
\newblock {\em arXiv preprint arXiv:1909.10094}, 2019.

\bibitem[\protect\citeauthoryear{Hasan \bgroup \em et al.\egroup }{2021}]{hasan2021xl}
Tahmid Hasan, Abhik Bhattacharjee, Md~Saiful Islam, Kazi Samin, Yuan-Fang Li, Yong-Bin Kang, M~Sohel Rahman, and Rifat Shahriyar.
\newblock Xl-sum: Large-scale multilingual abstractive summarization for 44 languages.
\newblock {\em arXiv preprint arXiv:2106.13822}, 2021.

\bibitem[\protect\citeauthoryear{Hayou \bgroup \em et al.\egroup }{2024}]{hayou2024lora+}
Soufiane Hayou, Nikhil Ghosh, and Bin Yu.
\newblock Lora+: Efficient low rank adaptation of large models.
\newblock {\em arXiv preprint arXiv:2402.12354}, 2024.

\bibitem[\protect\citeauthoryear{He \bgroup \em et al.\egroup }{2022}]{he2022kg}
Yong He, Cheng Wang, Shun Zhang, Nan Li, Zhaorong Li, and Zhenyu Zeng.
\newblock Kg-mtt-bert: Knowledge graph enhanced bert for multi-type medical text classification.
\newblock {\em arXiv preprint arXiv:2210.03970}, 2022.

\bibitem[\protect\citeauthoryear{He \bgroup \em et al.\egroup }{2025}]{DBLP:journals/inffus/HeMLRLFC25}
Kai He, Rui Mao, Qika Lin, Yucheng Ruan, Xiang Lan, Mengling Feng, and Erik Cambria.
\newblock A survey of large language models for healthcare: from data, technology, and applications to accountability and ethics.
\newblock {\em Information Fusion}, 118:102963, 2025.

\bibitem[\protect\citeauthoryear{Hettiarachchi \bgroup \em et al.\egroup }{2023}]{hettiarachchi2023ttl}
Hansi Hettiarachchi, Mariam Adedoyin-Olowe, Jagdev Bhogal, and Mohamed~Medhat Gaber.
\newblock Ttl: transformer-based two-phase transfer learning for cross-lingual news event detection.
\newblock {\em International Journal of Machine Learning and Cybernetics}, 2023.

\bibitem[\protect\citeauthoryear{Hu \bgroup \em et al.\egroup }{2020}]{hu2020ocnli}
Hai Hu, Kyle Richardson, Liang Xu, Lu~Li, Sandra K{\"u}bler, and Lawrence~S Moss.
\newblock Ocnli: Original chinese natural language inference.
\newblock {\em arXiv preprint arXiv:2010.05444}, 2020.

\bibitem[\protect\citeauthoryear{Huang \bgroup \em et al.\egroup }{2020}]{huang2020biomedical}
Kung-Hsiang Huang, Mu~Yang, and Nanyun Peng.
\newblock Biomedical event extraction with hierarchical knowledge graphs.
\newblock {\em arXiv preprint arXiv:2009.09335}, 2020.

\bibitem[\protect\citeauthoryear{Krause \bgroup \em et al.\egroup }{2022}]{krause2022generalized}
Franz Krause, Tobias Weller, and Heiko Paulheim.
\newblock On a generalized framework for time-aware knowledge graphs.
\newblock In {\em Towards a Knowledge-Aware AI}, pages 69--74. IOS Press, 2022.

\bibitem[\protect\citeauthoryear{Lai \bgroup \em et al.\egroup }{2023}]{lai2023towards}
Vivian Lai, Chacha Chen, Alison Smith-Renner, Q~Vera Liao, and Chenhao Tan.
\newblock Towards a science of human-ai decision making: An overview of design space in empirical human-subject studies.
\newblock In {\em Proceedings of the 2023 ACM Conference on Fairness, Accountability, and Transparency}, pages 1369--1385, 2023.

\bibitem[\protect\citeauthoryear{Li \bgroup \em et al.\egroup }{2021}]{li2021document}
Sha Li, Heng Ji, and Jiawei Han.
\newblock Document-level event argument extraction by conditional generation.
\newblock {\em arXiv preprint arXiv:2104.05919}, 2021.

\bibitem[\protect\citeauthoryear{Lin \bgroup \em et al.\egroup }{2023}]{DBLP:conf/acl/LinL0XC23}
Qika Lin, Jun Liu, Rui Mao, Fangzhi Xu, and Erik Cambria.
\newblock {TECHS:} temporal logical graph networks for explainable extrapolation reasoning.
\newblock In {\em Proceedings of the 61st Annual Meeting of the Association for Computational Linguistics (ACL)}, pages 1281--1293, 2023.

\bibitem[\protect\citeauthoryear{Lin \bgroup \em et al.\egroup }{2025a}]{DBLP:journals/corr/abs-2501-18119}
Qika Lin, Tianzhe Zhao, Kai He, Zhen Peng, Fangzhi Xu, Ling Huang, Jingying Ma, and Mengling Feng.
\newblock Self-supervised quantized representation for seamlessly integrating knowledge graphs with large language models.
\newblock {\em CoRR}, abs/2501.18119, 2025.

\bibitem[\protect\citeauthoryear{Lin \bgroup \em et al.\egroup }{2025b}]{DBLP:journals/inffus/LinZMHMHPCF25}
Qika Lin, Yifan Zhu, Xin Mei, Ling Huang, Jingying Ma, Kai He, Zhen Peng, Erik Cambria, and Mengling Feng.
\newblock Has multimodal learning delivered universal intelligence in healthcare? {A} comprehensive survey.
\newblock {\em Information Fusion}, 116:102795, 2025.

\bibitem[\protect\citeauthoryear{Ma \bgroup \em et al.\egroup }{2023}]{ma2023dreeam}
Youmi Ma, An~Wang, and Naoaki Okazaki.
\newblock Dreeam: Guiding attention with evidence for improving document-level relation extraction.
\newblock In {\em Proceedings of the 17th Conference of the European Chapter of the Association for Computational Linguistics}, pages 1971--1983, 2023.

\bibitem[\protect\citeauthoryear{Mirza and Tonelli}{2016}]{mirza2016catena}
Paramita Mirza and Sara Tonelli.
\newblock Catena: Causal and temporal relation extraction from natural language texts.
\newblock In {\em The 26th international conference on computational linguistics}, pages 64--75. ACL, 2016.

\bibitem[\protect\citeauthoryear{Ning \bgroup \em et al.\egroup }{2019}]{ning2019improved}
Qiang Ning, Sanjay Subramanian, and Dan Roth.
\newblock An improved neural baseline for temporal relation extraction.
\newblock {\em arXiv preprint arXiv:1909.00429}, 2019.

\bibitem[\protect\citeauthoryear{Paulus}{2017}]{paulus2017deep}
R~Paulus.
\newblock A deep reinforced model for abstractive summarization.
\newblock {\em arXiv preprint arXiv:1705.04304}, 2017.

\bibitem[\protect\citeauthoryear{Peng \bgroup \em et al.\egroup }{2023}]{peng2023knowledge}
Ciyuan Peng, Feng Xia, Mehdi Naseriparsa, and Francesco Osborne.
\newblock Knowledge graphs: Opportunities and challenges.
\newblock {\em Artificial Intelligence Review}, 56(11):13071--13102, 2023.

\bibitem[\protect\citeauthoryear{Pimenov \bgroup \em et al.\egroup }{2023}]{pimenov2023artificial}
Danil~Yu Pimenov, Andres Bustillo, Szymon Wojciechowski, Vishal~S Sharma, Munish~K Gupta, and Mustafa Kunto{\u{g}}lu.
\newblock Artificial intelligence systems for tool condition monitoring in machining: Analysis and critical review.
\newblock {\em Journal of Intelligent Manufacturing}, 34(5):2079--2121, 2023.

\bibitem[\protect\citeauthoryear{Sang and De~Meulder}{2003}]{sang2003introduction}
Erik~F Sang and Fien De~Meulder.
\newblock Introduction to the conll-2003 shared task: Language-independent named entity recognition.
\newblock {\em arXiv preprint cs/0306050}, 2003.

\bibitem[\protect\citeauthoryear{Savelka \bgroup \em et al.\egroup }{2023}]{savelka2023can}
Jaromir Savelka, Kevin~D Ashley, Morgan~A Gray, Hannes Westermann, and Huihui Xu.
\newblock Can gpt-4 support analysis of textual data in tasks requiring highly specialized domain expertise?
\newblock {\em arXiv preprint arXiv:2306.13906}, 2023.

\bibitem[\protect\citeauthoryear{Sui \bgroup \em et al.\egroup }{2023}]{sui2023joint}
Dianbo Sui, Xiangrong Zeng, Yubo Chen, Kang Liu, and Jun Zhao.
\newblock Joint entity and relation extraction with set prediction networks.
\newblock {\em IEEE Transactions on Neural Networks and Learning Systems}, 2023.

\bibitem[\protect\citeauthoryear{Wadhwa \bgroup \em et al.\egroup }{2023}]{wadhwa2023revisiting}
Somin Wadhwa, Silvio Amir, and Byron~C Wallace.
\newblock Revisiting relation extraction in the era of large language models.
\newblock In {\em Proceedings of the conference. Association for Computational Linguistics. Meeting}, volume 2023, page 15566. NIH Public Access, 2023.

\bibitem[\protect\citeauthoryear{Wang \bgroup \em et al.\egroup }{2021}]{wang2021survey}
Xin Wang, Yudong Chen, and Wenwu Zhu.
\newblock A survey on curriculum learning.
\newblock {\em IEEE transactions on pattern analysis and machine intelligence}, 44(9):4555--4576, 2021.

\bibitem[\protect\citeauthoryear{Wang \bgroup \em et al.\egroup }{2022}]{wang2022maven}
Xiaozhi Wang, Yulin Chen, Ning Ding, Hao Peng, Zimu Wang, Yankai Lin, Xu~Han, Lei Hou, Juanzi Li, Zhiyuan Liu, et~al.
\newblock Maven-ere: A unified large-scale dataset for event coreference, temporal, causal, and subevent relation extraction.
\newblock {\em arXiv preprint arXiv:2211.07342}, 2022.

\bibitem[\protect\citeauthoryear{Xu \bgroup \em et al.\egroup }{2024}]{xu2023symbol}
Fangzhi Xu, Zhiyong Wu, Qiushi Sun, Siyu Ren, Fei Yuan, Shuai Yuan, Qika Lin, Yu~Qiao, and Jun Liu.
\newblock Symbol-llm: Towards foundational symbol-centric interface for large language models.
\newblock In {\em ACL}, 2024.

\bibitem[\protect\citeauthoryear{Xu \bgroup \em et al.\egroup }{2025}]{xu2025large}
Fangzhi Xu, Qika Lin, Jiawei Han, Tianzhe Zhao, Jun Liu, and Erik Cambria.
\newblock Are large language models really good logical reasoners? a comprehensive evaluation and beyond.
\newblock {\em IEEE Transactions on Knowledge and Data Engineering}, 2025.

\bibitem[\protect\citeauthoryear{Yamada and Shindo}{2019}]{yamada2019neural}
Ikuya Yamada and Hiroyuki Shindo.
\newblock Neural attentive bag-of-entities model for text classification.
\newblock {\em arXiv preprint arXiv:1909.01259}, 2019.

\bibitem[\protect\citeauthoryear{Yao \bgroup \em et al.\egroup }{2019}]{yao2019docred}
Yuan Yao, Deming Ye, Peng Li, Xu~Han, Yankai Lin, Zhenghao Liu, Zhiyuan Liu, Lixin Huang, Jie Zhou, and Maosong Sun.
\newblock Docred: A large-scale document-level relation extraction dataset.
\newblock {\em arXiv preprint arXiv:1906.06127}, 2019.

\bibitem[\protect\citeauthoryear{Zhang \bgroup \em et al.\egroup }{2023}]{zhang2023instruction}
Shengyu Zhang, Linfeng Dong, Xiaoya Li, Sen Zhang, Xiaofei Sun, Shuhe Wang, Jiwei Li, Runyi Hu, Tianwei Zhang, Fei Wu, et~al.
\newblock Instruction tuning for large language models: A survey.
\newblock {\em arXiv preprint arXiv:2308.10792}, 2023.

\bibitem[\protect\citeauthoryear{Zhang \bgroup \em et al.\egroup }{2024}]{zhang2024semantic}
Jian Zhang, Changlin Yang, Haiping Zhu, Qika Lin, Fangzhi Xu, and Jun Liu.
\newblock A semantic mention graph augmented model for document-level event argument extraction.
\newblock In {\em Proceedings of the 2024 Joint International Conference on Computational Linguistics, Language Resources and Evaluation (LREC-COLING 2024)}, pages 1577--1587, 2024.

\bibitem[\protect\citeauthoryear{Zheng \bgroup \em et al.\egroup }{2023}]{zheng2023can}
Mingkai Zheng, Xiu Su, Shan You, Fei Wang, Chen Qian, Chang Xu, and Samuel Albanie.
\newblock Can gpt-4 perform neural architecture search?
\newblock {\em arXiv preprint arXiv:2304.10970}, 2023.

\end{thebibliography}

\appendix

\hspace{2cm}

\section{Details of Data Collection}
\label{DataCollection}

This section provides detailed information on all datasets of $\sim$806K pieces for training and $\sim$140K pieces for testing, including an overall introduction in Section~\ref{A1}, and the categorization of datasets into three types in Section~\ref{A2}.

\subsection{General Introduction} \label{A1}

As shown in Table~\ref{dataDetails}, we have collected, to the best of our ability, three types of different graph construction sub-task datasets for the GKG Dataset, along with an additional counter task (NLG task) dataset, resulting in a total of 15 sub-tasks and 29 datasets. To ensure data balance and reasonable distribution, we sample and partition some of the datasets. These sampling and partitioning processes are clearly indicated in Table~\ref{dataDetails} under the "Sampled?" field, allowing readers to better understand the data handling approach.

In the KG sub-task dataset, the focus is primarily on various types of relation extraction, including sentence-level relation extraction, few-shot relation extraction, and entity relation extraction, etc. This is because nodes in the KG sub-task are entities, and an important sub-task is to extract relationships between these entities. Furthermore, the EKG sub-task dataset primarily includes event detection, event argument extraction, and event relation extraction, as the event nodes are more complex, containing trigger words and various arguments. For the CKG sub-task dataset, the focus is closer to common-sense nodes and relations reasoning, involving tasks such as abstract generation and language inference.

\subsection{Three Categorizations} \label{A2}

The GKG Dataset is divided into three types: \textbf{in-domain data}, \textbf{counter task data}, and \textbf{OOD data}. The OOD data is separately indicated in Table~\ref{dataDetails} and is used only during the testing phase, not during training, to evaluate the model's performance on OOD data. The counter task is included to prevent overfitting and to enhance the generalizability of GKG-LLM.

Specifically, in-domain data consists of various GKG sub-tasks, combined with the counter task dataset (WebNLG) to form the training set. Using a curriculum learning fine-tuning framework, we obtained the final version of GKG-LLM. After testing on all in-domain datasets and the counter task dataset, we proceeded to test on three OOD datasets—TCR, Causal-TB, and R8—to validate the model's superior performance.

\begin{table*}[tp]
\centering
\footnotesize
\resizebox{\linewidth}{!}{
\begin{tabular}{lcc|cccccc}
    \toprule
    \textbf{Graphs} & \textbf{Tasks} &\textbf{Datasets} & \textbf{\# Train} & \textbf{\# Test} & \textbf{sampled?} & \textbf{\textit{held-out?}} &\textbf{Original Source}\\
    \midrule
    \multirow{6}{*}{\textbf{KG}} & SRE & NYT & 96,229 & 8,110 & & & \cite{paulus2017deep} \\
    & \multirow{2}{*}{FRE} & FewRel & 56,576 & 11,775 & & & \cite{han2018fewrel} \\
    & & TACRED & 18,448 & 3,325 & & & \cite{alt2020tacred} \\
    & DRE & DOCRED & 61,380 & 6,137 & \checkmark & & \cite{yao2019docred} \\
    & \multirow{2}{*}{JE\&RE} & FewRel & 28,288 & 11,775 & \checkmark & & \\
    & & NYT & 48,114 & 8,110 & \checkmark & & \\
    \midrule
    \multirow{15}{*}{\textbf{EKG}} & SED & ACE2005 & 3,681 & 409 & & & \cite{grishman2005nyu} \\
    & DED & WIKIEVENTS & 3,586 & 365 & & & \cite{li2021document} \\
    & \multirow{2}{*}{DEAE} & WIKIEVENTS & 3,586 & 365 & & & \\
    &  & RAMS & 7,339 & 761 & & & \cite{ebner2020multi} \\
    & \multirow{6}{*}{ETRE} & MATRES & 12,216 & 1,361 & & & \cite{ning2019improved}\\
    & & ESL & 7,652 & 852 & & & \\
    & & TB-Dense & 9,257 & 2,639 & & & \cite{han2019deep}\\
    & & Causal-TB & 5,427 & 603 & & & \cite{mirza2016catena}\\
    & & MAVEN-ERE & 80,000 & 5,000 & \checkmark & & \cite{wang2022maven}\\
    & & TCR & & 3,515 & & \checkmark & \cite{han2019deep}\\
    & \multirow{3}{*}{ECRE} & ESL & 3,196 & 356 & & & \\
    & & MAVEN-ERE & 63,980 & 7,330 & \checkmark & & \\
    & & Causal-TB & & 318 & & \checkmark & \\
    & \multirow{2}{*}{ESRE} & HiEve & 12,107 & 1,348 & & & \cite{glavavs2014hieve}\\
    & & MAVEN-ERE & 31,365 & 4,244 & & & \\
    \midrule
    \multirow{7}{*}{\textbf{CKG}} & NER & CoNLL & 17,293 & 3,454 & & & \cite{sang2003introduction} \\
    & \multirow{2}{*}{AG} & CNNDM & 51,684 & 11,490 & \checkmark & & \cite{chen2021dialogsum} \\
    & & XSum & 50,666 & 11,334 & \checkmark & & \cite{hasan2021xl} \\
    & \multirow{2}{*}{LI} & SNLI & 50,000 & 10,000 & \checkmark & & \cite{camburu2018snli} \\
    & & MNLI & 50,000 & 10,000 & \checkmark & & \cite{hu2020ocnli} \\
    & \multirow{2}{*}{TC} & R8 & & 7,674 & & \checkmark & \cite{yamada2019neural} \\
    & & R52 & 7,816 & 1,284 & \checkmark & & \cite{ge2017improving} \\
    \midrule
    \textit{Counter} & NLG & WebNLG & 26,302 & 6,513 & & & \cite{gardent2017webnlg} \\
    \bottomrule
\end{tabular}}
\caption{Detailed illustrations of 15 sub-task types across 29 datasets, categorized within three types of graphs, along with a counter dataset—WebNLG. \# Train and \# Test represent the number of training and testing samples, respectively. \textit{Sampled?} indicates whether the dataset is sampled from the original to achieve data balancing. \textit{Held-out?} specifies whether the dataset is used during the training phase. \textit{Original Source} refers to the citation of the original paper.}
\label{dataDetails}
\end{table*}

\section{Data Format}
\label{DataFormat}

\begin{figure}[t]
    \includegraphics[width=\columnwidth]{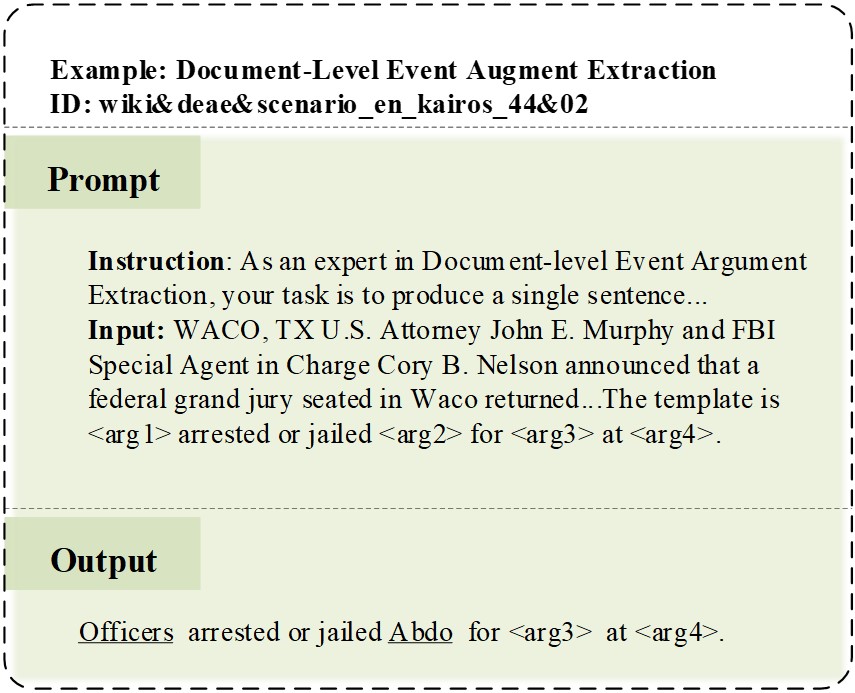}
    \caption{An example from the WIKEVENTS dataset. It consists of five fields \(ID\), instruction \( s_i \), few-shot \( fs\) / zero-shot \( zs\) , input \(x_i\), and output \(y_i\).}
    \label{data_format}
\end{figure}

To bridge the gap between the dataset's data format and the instruction-tuning format, we formatted all the data. Specifically, each data entry consists of five fields-- \(ID\), instruction \( s_i \), few-shot \( fs\) / zero-shot \( zs\) , input \(x_i\), and output \(y_i\).  as shown in Figure~\ref{data_format}, this example is from the WIKIEVENTS dataset. \(ID\) represents the unique identifier of each data entry, which includes the task name, dataset name, and specific data entry. The instruction \( s_i \) provides a formal definition of each sub-task and is passed to the base model to help it understand the task's intent. few-shot \( fs\) / zero-shot \( zs\) field indicates whether a few-shot example is included in the prompt; in particular, for zero-shot, this field can be omitted. The input \(x_i\)represents the specific input data, while the output \(y_i\) represents the corresponding output.

To more comprehensively simulate real-world scenarios, we utilize GPT-4 to generate ten \textbf{diverse instructions}, which are then randomly assigned to the instruction field of each data entry. This approach aims to enhance the model's ability to understand and handle a variety of task instructions, thereby increasing its flexibility and adaptability for real-world multitasking needs. By diversifying the instructions, we aim to train the model to better respond to different directives, similar to a practical deployment setting. Additionally, for 10\% of the data pieces, we randomly added a \textbf{few-shot example} to help the base model understand the task structure more effectively. The majority of the data entries, however, remained in a zero-shot setting, ensuring that the model could learn general patterns of GKG construction tasks without extensive direct guidance. By balancing few-shot and zero-shot learning, we aim to improve the model's generalization capabilities across a range of GKG-related tasks.

\section{Stage Generalization} \label{3.3}

\begin{figure}[h]
    \centering
    \includegraphics[width=\columnwidth]{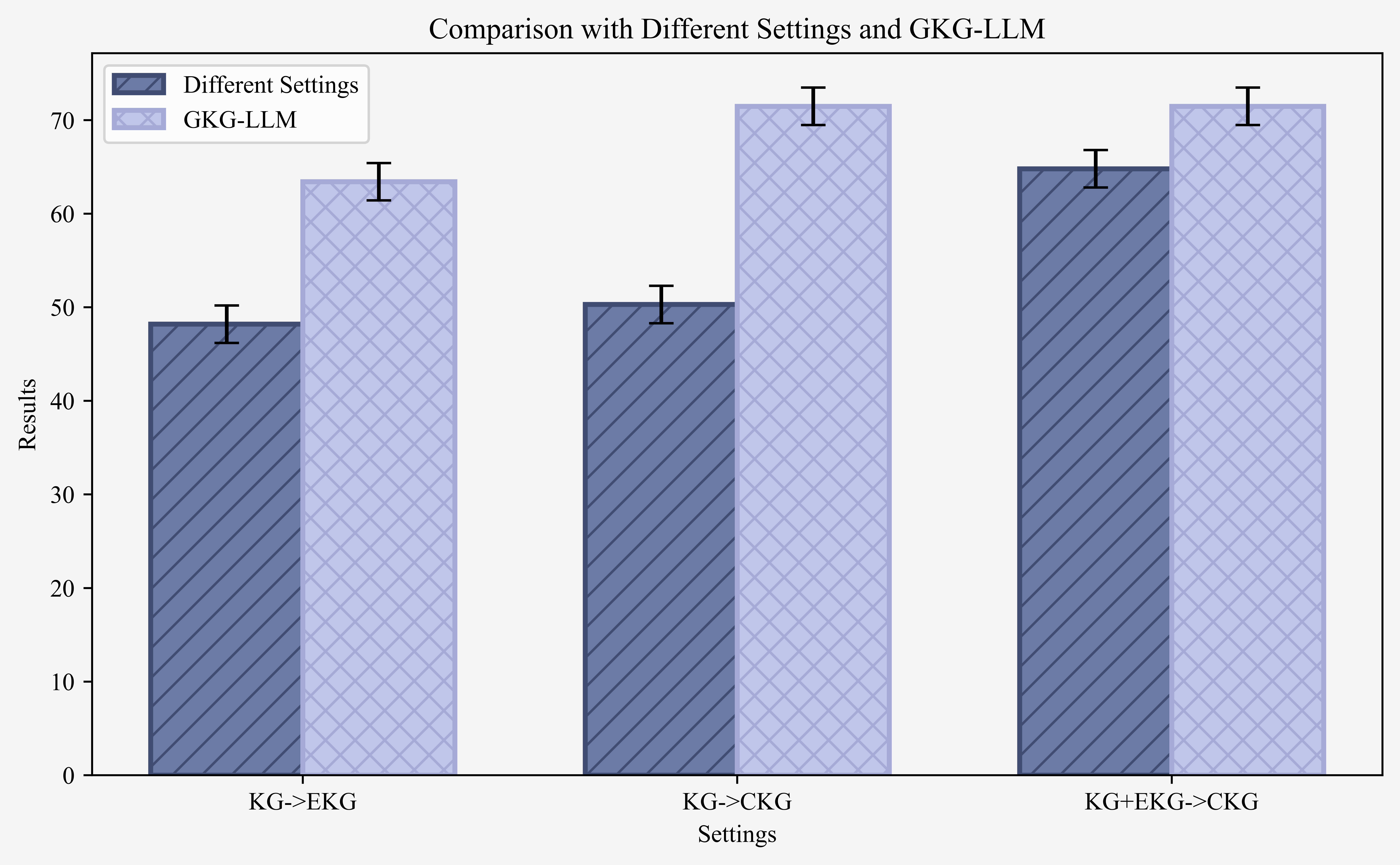}
    \caption{Comparison of Results by different settings and GKG-LLM.}
    \label{fig:individual_data_vs_all_data}
\end{figure}

In this section, we examine the effect of the three-stage training strategy on subsequent data exploration stages. Specifically, we test G-Micro, trained only on KG-related sub-task datasets, on EKG and CKG sub-task datasets, and G-Mid on the CKG sub-task dataset. The results are shown in Figure \ref{fig:individual_data_vs_all_data}.

The experimental results show that, despite some trade-offs in the exploratory experiments, the three-stage curriculum learning approach achieves superior performance. This demonstrates: \textbf{1).} earlier GKG-LLM versions influence subsequent tasks, indicating task correlation; \textbf{2).} the unified approach to the three types of graphs in GKG is valuable and meaningful, reflecting their progressive relationship within a unified framework.



\section{Exploration of LoRA+ Hyperparameter Values} \label{hyperparameters}

As described in Section 2.3, we adopt the LoRA+ training strategy, where the low-rank matrices \(A\) and \(B\) have different rates of change, meaning they each have distinct hyperparameters \(\eta_A\) and \(\eta_B\).

\begin{figure}[h]
    \centering
    \includegraphics[width=\linewidth]{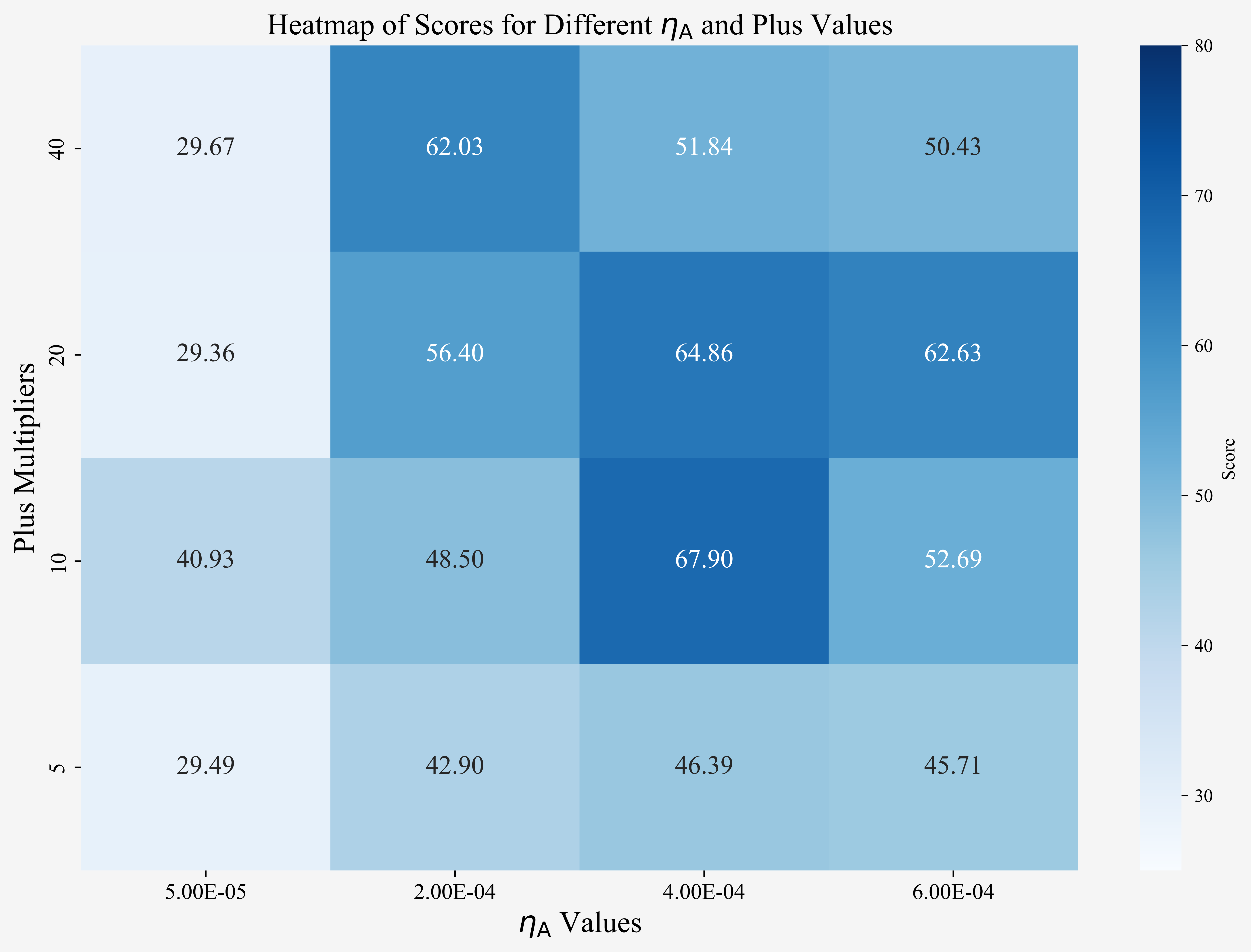}
    \caption{Heatmap of Scores for Different \(\eta_A\) and \(\eta_B\) Values for our training strategy.}
    \label{fig:lora_hyperparameter_exploration}
\end{figure}

In this section, we explore the effects of different combinations of the hyperparameters \(\eta_A\) and \(\eta_B\) on the model's performance. The experimental results are illustrated in Figure \ref{fig:lora_hyperparameter_exploration}, the vertical axis represents \(B\), which is expressed as a multiple of \(\eta_A\). The model's performance is highly sensitive to changes in \(\eta_A\) and \(\eta_B\). The highest performance score of 67.90\% was achieved with \(\eta_A = 4 \times 10^{-4}\) and \(\eta_B = 4 \times 10^{-3}\). This suggests that higher learning rates for \(\eta_A\) combined with moderate values of \(\eta_B\) are beneficial for fine-tuning. Conversely, the lowest performance scores were observed with the smallest value of \(\eta_A = 5 \times 10^{-5}\), regardless of the value of \(\eta_B\). This indicates that too low a learning rate for the adaptation matrices may not be sufficient for effective fine-tuning. Increasing \(\eta_B\) tends to enhance performance up to a certain point, after which the performance gains stabilize or diminish. For example, \(\eta_A = 2 \times 10^{-4}\) with \(\eta_B = 8 \times 10^{-3}\) shows a obvious score, but further increasing \(\eta_B\) does not yield substantial improvements.

These findings highlight the importance of carefully tuning the hyperparameters \(\eta_A\) and \(\eta_B\) in the LoRA+ framework to achieve optimal model performance. The insights gained from this exploration can guide future experiments and the development of more effective fine-tuning strategies for LLMs. In summary, the exploration of LoRA+ hyperparameters reveals that selecting the appropriate values for \(\eta_A\) and \(\eta_B\) is crucial for maximizing model performance. This study provides a foundational understanding that can be leveraged to further enhance the efficiency and effectiveness of fine-tuning LLMs using low-rank adaptation techniques.

\section{Hyper-parameters} \label{stages}

In the implementation, we leverage the LoRA+ technique to fine-tune models using four A800 (80GB) GPUs, with a maximum sequence length of 4,096. The fine-tuning process is optimized with FlashAttention2, while the AdamW optimizer is employed with a learning rate of 5e-5 across three curriculum learning stages, each controlled by a linear learning rate scheduler. We use one epoch per stage to complete the tuning process.

During the KG empowerment stage, model weights are initialized from LLaMA-3-Instruct, resulting in the tuned model named G-Micro. In the EKG enhancement stage, G-Micro serves as the starting point, producing G-Mid. Similarly, in the CKG generalization stage, we initialize from G-Mid and ultimately obtain GKG-LLM. Inference process is conduct on a single A800 (80GB) GPU using greedy search.

\section{Sub-tasks Introduction}
\label{sub-tasks}
The GKG dataset is composed of three types of sub-task datasets: KG , EKG and CKG. The data is categorized into three types: In-domain data, OOD data, and counter-task data. The specific descriptions of these tasks are as follows.

\subsection{KG} \label{C1}

\paragraph{SRE (Sentence-level Relation Extraction)}
For the SRE task, we utilize the NYT dataset. This task focuses on identifying the entities mentioned in a complex news sentence and, based on entity recognition, detecting and labeling the relationships between the entities. This task plays a critical role in the process of transforming unstructured textual data into structured knowledge.

\paragraph{FRE (Few-shot Relation Extraction)}
Due to the issue of insufficient labeled corpora in many domains and the high cost of manual annotation, the FRE task aims to train a model using a small amount of labeled sample data, enabling the model to learn the characteristic information of entities that form relationships. During the testing phase, the model is asked to identify previously unseen relationship types from new datasets. In our work, we utilize the FewRel and TACRED datasets for both training and testing.

\paragraph{DRE (Document-level Relation Extraction)}
Compared to SRE, the DRE task is more challenging, as it requires the model not only to identify relations within a single sentence but also to understand the context and possess the ability to recognize relations across sentences and even across paragraphs. In this paper, we conduct experiments using the DocRED dataset. The input is a long text document containing multiple sentences and entities, while the output consists of all entity pairs in the document and their corresponding relation types.

\paragraph{JE\&RE (Entity-Relation Joint Extraction)}
The previously mentioned relation extraction approaches follow a pipeline where entity recognition is performed first, followed by relation classification based on the identified entities. In contrast, JE\&RE task differs by requiring the model to extract both entities and relations simultaneously, without dividing the process into two separate tasks. In this work, we conduct experiments using the FewRel and NYT datasets.

\subsection{EKG} \label{C2}

\paragraph{SED (Sentence-level Event Detection)}
Event detection (ED) aims to identify the events mentioned in a given text and recognize their characteristics, such as event type, participants, time, and other relevant attributes. SED is a specific form of ED, where the task requires the model to detect events within individual sentences. In this work, we utilize the ACE2005 dataset for training and testing the model.

\paragraph{DED (Document-level Event Detection)}
DED aims to identify multiple events within a document and extract relevant information, such as participants, triggers, and other attributes. Since these events may be distributed across different sentences, DED requires the model to have cross-sentence contextual understanding, making it more complex and enriched compared to sentence-level tasks. In this work, we use the WIKIEVENTS dataset, leveraging Wikipedia entries as events to train and test the model.

\paragraph{DEAE(Document-level Event Argument Extraction)}
 DEAE is a task designed to extract argumentative material from a full document, requiring the identification of arguments in a relationship and the extraction of the relations between arguments and events. In our work, we train and test the model using the WIKIEVENTS and RAMS datasets, where the RAMS dataset includes a rich set of argument types and deals with the relations of argument elements between different sentences.
 
\paragraph{ETRE (Event Temporal Relation Extraction)}
ETRE aims to extract events mentioned in a text and determine the temporal order in which these events occur. In our experiments, we use the MATRES, ESL, TB-Dense, Causal-TB, MAVEN-ERE, and TCR datasets for training and testing the model. Notably, the TCR dataset, as an \textbf{OOD dataset}, is only used for testing and not for training.

\paragraph{ECRE (Event Causal Relation Extraction)}
ECRE aims to identify and extract causal relationships between different events in a text. In our work, we use the ESL and MAVEN-ERE datasets for training and testing the model. The ESL dataset is further annotated with various types of causal relationships between events, including direct causality, indirect causality, and opposition relationships. Additionally, during testing, we employ the Causal-TB dataset as an \textbf{OOD dataset}, which is only used for testing and not for training.

\paragraph{ESRE (Event Subevent Relation Extraction)}
In complex texts, events often do not exist independently but can exhibit hierarchical structures, where one event may be the cause, effect, or sub-event of another. ESRE aims to identify these hierarchical relationships between events to achieve a more comprehensive understanding of the event timeline and causal chains. The input to this task is typically a text containing multiple events, and the output is pairs of events along with their hierarchical relationship labels, such as parent event and child event, causal relation, and parallel relation. In this work, we use the HiEve and MAVEN-ERE datasets for model training and testing.

\subsection{CKG} \label{C3}
\paragraph{NER (Named Entity Recognition)}
NER aims to identify entities with specific semantic meanings from a text and classify them into predefined categories, such as person names, locations, organizations, dates, times, and numerical values. Given a natural language text as input, the output consists of the extracted named entities and their corresponding categories. NER plays a critical role in the construction of knowledge graphs by recognizing entities in the text and linking them to existing entity nodes in the knowledge graph, facilitating the automated development and expansion of the graph. In this work, we use the CoNLL dataset for training and testing the NER task.

\paragraph{AG (Abstract Generation)}
AG aims to compress a lengthy input text into a concise and accurate abstract while retaining key information and themes. Since CKG can provide rich background and relational information, we employ a CKG-based abstraction task. For this purpose, we train and test the model using the CNNDM and XSum datasets, with the \textbf{ROUGE-L} percentage metric used as the evaluation criterion.

\paragraph{LI (Language Inference)}
The task of LI aims to establish an understanding of relationships between sentences. The core objective of this task is to determine whether a given pair of sentences exhibits entailment, contradiction, or neutrality. Typically, the input consists of a pair of texts, and the output indicates whether the relationship between the two sentences is entailment, contradiction, or neutral. In this work, we use two specialized datasets in the field of natural language inference, the SNLI and MNLI datasets, for training and testing the model.

\paragraph{TC (Text Classification)}
TC task aims to automatically assign textual data to one or more predefined categories. Given a text as input, the output is typically the predicted category or categories corresponding to the input text. In this work, we use the R8 and R52 datasets for model training and testing, with R8 serving as an \textbf{OOD dataset} that is used only for training and not for testing.

\subsection{Counter} 
\paragraph{NLG (Natural Language Generation)}
NLG aims to generate natural language text in a predefined format or structure based on specific input information or structure. Unlike traditional free-text generation, the structured text generation task emphasizes the structure and accuracy of the information in the output. The input can take various forms of structured data, such as knowledge graphs, tables, or tuples, and the output is typically a coherent piece of text that adheres to the predetermined structure. In this work, we use the WebNLG dataset, a typical dataset in this domain, for model training and testing. Specifically, we employ the \textbf{ROUGE-L} percentage metric as the evaluation criterion.

\end{document}